\documentclass[lettersize,journal]{IEEEtran}
\usepackage{stfloats}
\usepackage{url}
\usepackage{verbatim}
\usepackage{graphicx}
\usepackage{balance}
\usepackage{textcomp}
\usepackage{xcolor}

\usepackage{amsmath,amssymb,amsfonts}
\usepackage{bm}

\usepackage{booktabs}
\usepackage{multirow}
\usepackage{tabularx}

\usepackage[caption=false,font=normalsize,labelfont=sf,textfont=sf]{subfig}

\usepackage[hypertexnames=false,colorlinks=true,allcolors=blue,linktoc=all]{hyperref}

\usepackage[capitalise]{cleveref}

\usepackage{setspace}

\def\BibTeX{{\rm B\kern-.05em{\sc i\kern-.025em b}\kern-.08em
    T\kern-.1667em\lower.7ex\hbox{E}\kern-.125emX}}

\begin{document}
\vspace{0.5em}
\title{Towards Privacy-Preserving Mental Health Support with Large Language Models}
\author{Dong~Xue,~\IEEEmembership{Member,~IEEE}, Jicheng~Tu, Ming~Wang, Xin~Yan, Fangzhou~Liu,~\IEEEmembership{Member,~IEEE} and Jie~Hu 
\thanks{This work has been submitted to the IEEE for possible publication. Copyright may be transferred without notice, after which this version may no longer be accessible.}
\thanks{Corresponding author: Dong Xue.}}
\maketitle

\begin{abstract}
Large language models (LLMs) have shown promise for mental health support, yet training such models is constrained by the scarcity and sensitivity of real counseling dialogues. In this article, we present MindChat, a privacy-preserving LLM for mental health support, together with MindCorpus, a synthetic multi-turn counseling dataset constructed via a multi-agent role-playing framework. To synthesize high-quality counseling data, the developed dialogue-construction framework employs a dual closed-loop feedback design to integrate psychological expertise and counseling techniques through role-playing: (i) turn-level critique-and-revision to improve coherence and counseling appropriateness within a session, and (ii) session-level strategy refinement to progressively enrich counselor behaviors across sessions. To mitigate privacy risks under decentralized data ownership, we fine-tune the base model using federated learning with parameter-efficient LoRA adapters and incorporate differentially private optimization to reduce membership and memorization risks. Experiments on synthetic-data quality assessment and counseling capability evaluation show that MindCorpus improves training effectiveness and that MindChat is competitive with existing general and counseling-oriented LLM baselines under both automatic LLM-judge and human evaluation protocols, while exhibiting reduced privacy leakage under membership inference attacks.
\end{abstract}


\section{Introduction}\label{section::I}
Mental health is a fundamental component of overall well-being and has become a growing public health priority. Globally, approximately $4.7\%$ of the population experiences mental health disorders such as depression each year~\cite{herrman2022time}. The incorporation of artificial intelligence (AI) into psychotherapeutic practices presents a viable solution for broadening access to mental health assistance. Specifically, AI applications in the field of psychology cover affective computing~\cite{hao2025interview}, disease diagnosis~\cite{11251119}, and therapeutic interventions~\cite{lee2021artificial}. Recent advances in large language models (LLMs), including ChatGPT~\cite{achiam2023gpt}, DeepSeek~\cite{liu-2024-deepseek}, and Qwen~\cite{yang2025qwen3}, has greatly broadened the range of practical AI applications. However, the responses generated by the general LLMs in the psychological counseling service scenario are often broad but lack depth and professional targeting. This limitation has prompted the emergence of a research route, that is, to improve model performance by fine-tuning on high-quality consulting datasets in specific fields. Representative efforts include SoulChat~\cite{chen-etal-2023-soulchat}, CPsyCounX~\cite{zhang-etal-2024-cpsycoun}, EmoLLM~\cite{yang2024emollm}, and MeChat~\cite{qiu-etal-2024-smile}. These efforts significantly improve the professionalism of the model in psychological counseling applications, and further emphasize the importance of high-quality emotional support datasets and effective fine-tuning strategies for the development of reliable and professional mental health LLMs.

Supervised fine-tuning of LLMs for psychological counseling typically requires large-scale multi-round dialogue datasets. However, authentic counseling conversations are scarce and often inaccessible due to privacy considerations, which greatly increases the difficulty of data collection~\cite{Liu_ACL_20}. To meet this challenge, prior work explores generating training data by prompting LLMs with seed instructions and partial dialogue fragments, leveraging their capacity to complete and expand consultative dialogue~\cite{zheng-etal-2023-augesc}. Techniques like cognitive restructuring~\cite{xiao-etal-2024-healme} and reasoning-augmented prompting~\cite{zhang-etal-2024-escot} are introduced to improve the authenticity of generated dialogues and their alignment with therapeutic principles. Nevertheless, most current approaches adopt a one-pass generation workflow, where data quality is ensured through subsequent filtering or manual revision, rather than iterative optimization through dynamic feedback mechanisms. Based on the constructed data, the training of LLMs typically relies on fine-tuning strategies, which can be generally categorized into full fine-tuning and parameter-efficient fine-tuning (PEFT). Full fine-tuning requires updating all model parameters during the training process, which puts forward considerable requirements in terms of data volume and computing resources. In contrast, PEFT only adjusts a small subset of parameters, which can still achieve competitive performance while significantly reducing the cost of training~\cite{ding2023parameter}. Representative PEFT methods include Low-Rank Adaptation (LoRA)~\cite{hu2022lora}, P-Tuning~\cite{liu-etal-2022-p}, and Adapter Tuning~\cite{chen2023hadamard}, among which LoRA gains widespread adoption for its favorable trade-off between computational efficiency and task-specific performance. 

Dialogues in psychological counseling often involve highly sensitive personal information, making their use for model training a source of significant ethical and security concerns. Additionally, such data are typically distributed across hospitals, counseling clinics, or online platforms, showing decentralization characteristics, while local institutions need to ensure that private records are not leaked externally. These constraints hinder centralized data aggregation. Addressing these challenges requires a training paradigm that supports decentralized collaboration with low communication overhead, alongside robust privacy protection mechanisms. Meanwhile, privacy risks related to LLMs have attracted growing attention in recent studies. For instance, the combination of jailbreaking attacks and chain-of-thought reasoning can induce models like ChatGPT to disclose confidential personal information~\cite{li-etal-2023-multi-step}. From a data security standpoint, incorporating differential privacy (DP) noise into the training process serves as an effective means of mitigating privacy breaches~\cite{YAO2024100211}. While existing studies explore privacy protection in general LLMs~\cite{wang2025unique} and federated learning (FL) is applied to train domain-specific LLMs on distributed data~\cite{cheng2024towards}, a psychological LLM that comprehensively addresses the privacy protection of training data remains under-explored.

In this article, a novel multi-agent collaborative architecture is introduced to generate high-quality psychological dialogue data through role-playing. To ensure the reliability and applicability of the synthesized dataset, a dual-loop dynamic feedback mechanism is integrated for iterative evaluation and refinement, resulting in \emph{MindCorpus}, a dataset comprising $5.7k$ counseling sessions. Moreover, five evaluation metrics are proposed to comprehensively assess the quality of the synthesized data. To address the critical issue of data privacy, a privacy-preserving fine-tuning approach is employed, combining the FL technique and DP mechanism. FL enables the training of a global model by aggregating locally trained models without centralizing sensitive data, while DP is incorporated during training to minimize the risk of exposing the underlying corpus. Furthermore, LoRA is adopted for local model optimization, substantially reducing computational overhead and communication costs, which improves the overall efficiency and scalability of the approach. Building upon \emph{MindCorpus} and the proposed training paradigm, an AI-powered psychological counseling assistant has been developed to deliver professional, empathetic, and privacy-preserving mental health support. The contributions of this work can be summarized as follows:
\begin{itemize}
    \item A multi-agent collaborative framework with a dual closed-loop feedback mechanism is proposed for synthesizing high-quality multi-round counseling dialogues, yielding \emph{MindCorpus}, a contextually rich dataset for mental health support scenarios. In addition, a comprehensive evaluation scheme comprising five dimensions is introduced to assess dialogue quality from both seeker and supporter perspectives.
    \item A privacy-preserving and efficient distributed fine-tuning paradigm for LLMs in psychological scenarios is established by integrating FL, DP, and LoRA. This framework enables collaborative model training without centralizing sensitive counseling data, while providing formal privacy guarantees and maintaining practical training efficiency.
    \item Experimental results demonstrate that \emph{MindCorpus} attains superior dialogues quality compared to existing emotional support datasets, and the trained chatbot \emph{MindChat} achieves competitive performance against both general-purpose and specialized psychological LLMs on key metrics, while emphasizing the privacy protection of training data.
\end{itemize}

\section{Related Works}\label{RW}
\subsection{Data Construction for LLMs}
The scarcity of high-quality data poses a major challenge to the advancement of LLMs. In order to alleviate this issue, data synthesis and augmentation emerge as effective strategies for enriching training corpus~\cite{chai2026text}. For instance, ChatGPT is employed to generate multi-round dialogues to improve the performance of models such as LLaMA~\cite{xu-etal-2023-baize}. In the medical domain, HuatuoGPT~\cite{zhang-etal-2023-huatuogpt} demonstrates that ChatGPT-distilled dialogues, when combined with real-world clinical data, can effectively support supervised fine-tuning of medical consultation LLMs. 

Similarly, in the field of mental health applications, researchers enrich datasets by synthesizing multi-turn counseling dialogues from limited sources. Early efforts focus on expanding single-turn data: SMILECHAT~\cite{qiu-etal-2024-smile} uses ChatGPT to rewrite queries into empathetic multi-turn communication via prompt-based extensions. On this basis, SoulChatCorpus~\cite{chen-etal-2023-soulchat} strengthens empathy constraints in prompting and applies manual proofreading to produce a large-scale Chinese dataset, emphasizing supportive behaviors such as active listening and emotional validation. Despite their scale and empathetic design, these approaches often lack grounding in professional counseling knowledge, which limits their therapeutic authenticity. Subsequent work seeks to ground synthetic dialogues in domain expertise. CPsyCounD~\cite{zhang-etal-2024-cpsycoun} constructs dialogues directly from psychological counseling reports using a two-phase Memo2Demo pipeline, explicitly embedding counseling principles into synthetic interactions. PsyDTCorpus~\cite{xie2025psydt} further explores this direction by leveraging dynamic one-shot learning with GPT-4 to capture counselor linguistic styles and therapy techniques, generating personalized multi-turn dialogues conditioned on client personality. More recently, domain-specific scenarios are explored. For instance, PeConv~\cite{zhao2025parental} targets parent–child emotional support by integrating child emotion coaching theory into a human–machine collaborative framework, and represents an early effort toward constructing a Chinese dialogue dataset for parental counseling assistance.

Nevertheless, most existing approaches rely on single-model generation or template-based transformations, often lacking dynamic interaction, role-specific expertise, and iterative quality improvement. As a result, the generated dialogues may lack authenticity and contextual depth in simulating real psychological counseling sessions. To bridge this gap, this paper proposes a novel multi-agent collaborative framework that employs a role-playing mechanism to simulate realistic therapeutic interactions, enabling the generation of high-quality, expert-informed mental health dialogues.

\subsection{Privacy-preserving LLMs for Mental Health} 
Recent studies explore adapting LLMs to mental health support scenarios through fine-tuning on domain-specific multi-turn dialogue data, as exemplified by SoulChat~\cite{chen-etal-2023-soulchat}, CPsyCounX~\cite{zhang-etal-2024-cpsycoun}, EmoLLM~\cite{yang2024emollm}, and MeChat~\cite{qiu-etal-2024-smile}. These works demonstrate that incorporating counseling-oriented interaction histories enhances empathy and response quality in psychological LLMs. Despite their effectiveness, these methods all use the centralized training mode. Such a prerequisite inevitably raises privacy concerns, as mental health data is highly sensitive and subject to strict regulations such as HIPAA and GDPR, which greatly restrict the applicability of these fine-tuning approaches in settings where data sharing across institutions or individuals is not permitted. Moreover, LLMs exhibit a tendency to memorize training data, resulting further risks of sensitive information being reconstructed through adversarial or model inversion attacks~\cite{Feldman-2020-memory}.

In order to mitigate these risks, existing efforts on privacy-preserving LLMs mainly focus on data preprocessing and model-level defenses. At the data level, anonymization techniques, such as removing or masking personal identification information, are commonly employed to reduce privacy exposure~\cite{zuo-2021-ano1}. Unlike anonymization, which may alter the original content of the data, model-level approaches preserve data integrity while providing privacy protection during training. Among these, FL serves as a promising paradigm for scenarios where sensitive data is distributed across multiple parties, enabling collaborative model training without centralizing raw data~\cite{chen-2024-fl}.

In mental health domain, FL has been explored for situations including facial emotion recognition~\cite{chatterjee2024federated} and depression detection from multilingual textual data~\cite{khalil2024federated}. These studies demonstrate the worth of FL in reducing data exposure risks, yet they primarily focus on classification tasks rather than generative language modeling. More recent work extends FL to LLMs for mental health applications. FedMentalCare~\cite{sarwar2025fedmentalcare} presents a federated framework with PEFT for mental health analysis, demonstrating scalability and reduced computational overhead. Nevertheless, this line of research does not incorporate formal privacy protection mechanisms, nor does it consider multi-turn counseling dialogues that are essential for modeling empathetic‌ interactions. To further defend against privacy leakage through model parameters, DP is often integrated into federated systems through central or local mechanisms~\cite{hu2024does}. Building on these insights, the present work introduces a privacy-preserving framework that combines FL with local DP to fine-tune psychological LLMs on multi-turn counseling dialogues, reducing the risk of data reconstruction while maintaining training efficiency in distributed mental health scenarios.

\section{Methodology}\label{method}
\begin{figure}[!ht]
\centering
    \includegraphics[width=1.0\columnwidth]{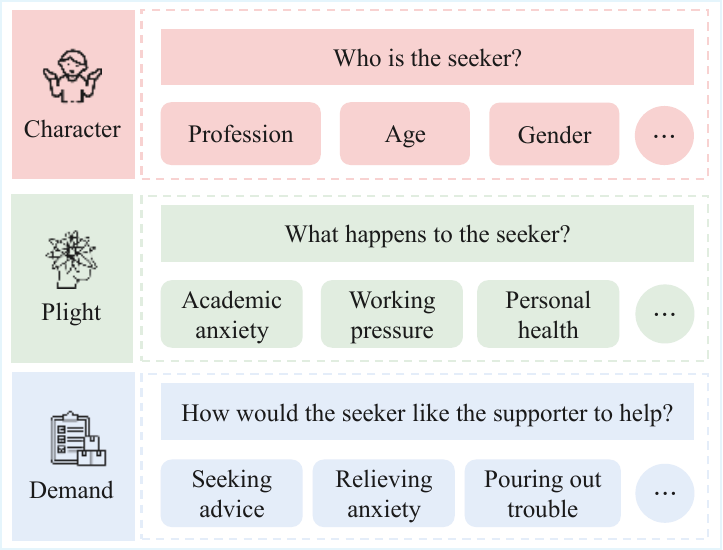}
    \caption{The personal situation of seekers.}
    \label{seeker_description}
\end{figure}
This section introduces the methodology underlying the proposed \emph{MindChat} framework. It begins with a description of mental health support data construction, including the sources and preprocessing of seed data and the novel dual-loop multi-agent data generation framework. Based on the resulting multi-round dialogue dataset for psychological counseling, the training framework of \emph{MindChat} is subsequently presented, covering parameter-efficient fine-tuning and federated learning with integrated privacy protection. Finally, the evaluation protocols for both the counseling datasets and the empathetic mental health LLM are outlined.
\subsection{Mental Health Support Data Construction}
\subsubsection{Seed data}
The situation of individuals seeking psychological support play a crucial role in generating realistic mental health dialogues. Motivated by~\cite{zhang-etal-2024-escot}, the seed dataset is constructed using approximately $11$k situation texts related to psychological seekers, collected from the publicly accessible online counseling platforms Yixinli and Jiandanxinli. The corpus covers a broad spectrum of psychological and behavioral areas, including emotional and relationship management, self-awareness and individual growth, stress and anxiety relief, mental health maintenance, and workplace adjustment. Collectively, these diverse and thematically comprehensive data provide a solid foundation for simulating real psychological counseling scenarios. However, the raw data exhibit gaps in key contextual information and often contain overly colloquial expressions. For this reason, we use the GLM-4-Plus model to clean and enrich the original materials. The resultant individual situation representation comprises three core elements: Character, Plight, and Demand, as illustrated in~\cref{seeker_description}. 

\begin{figure*}[!ht]
\centering
    \includegraphics[width=1.6\columnwidth]{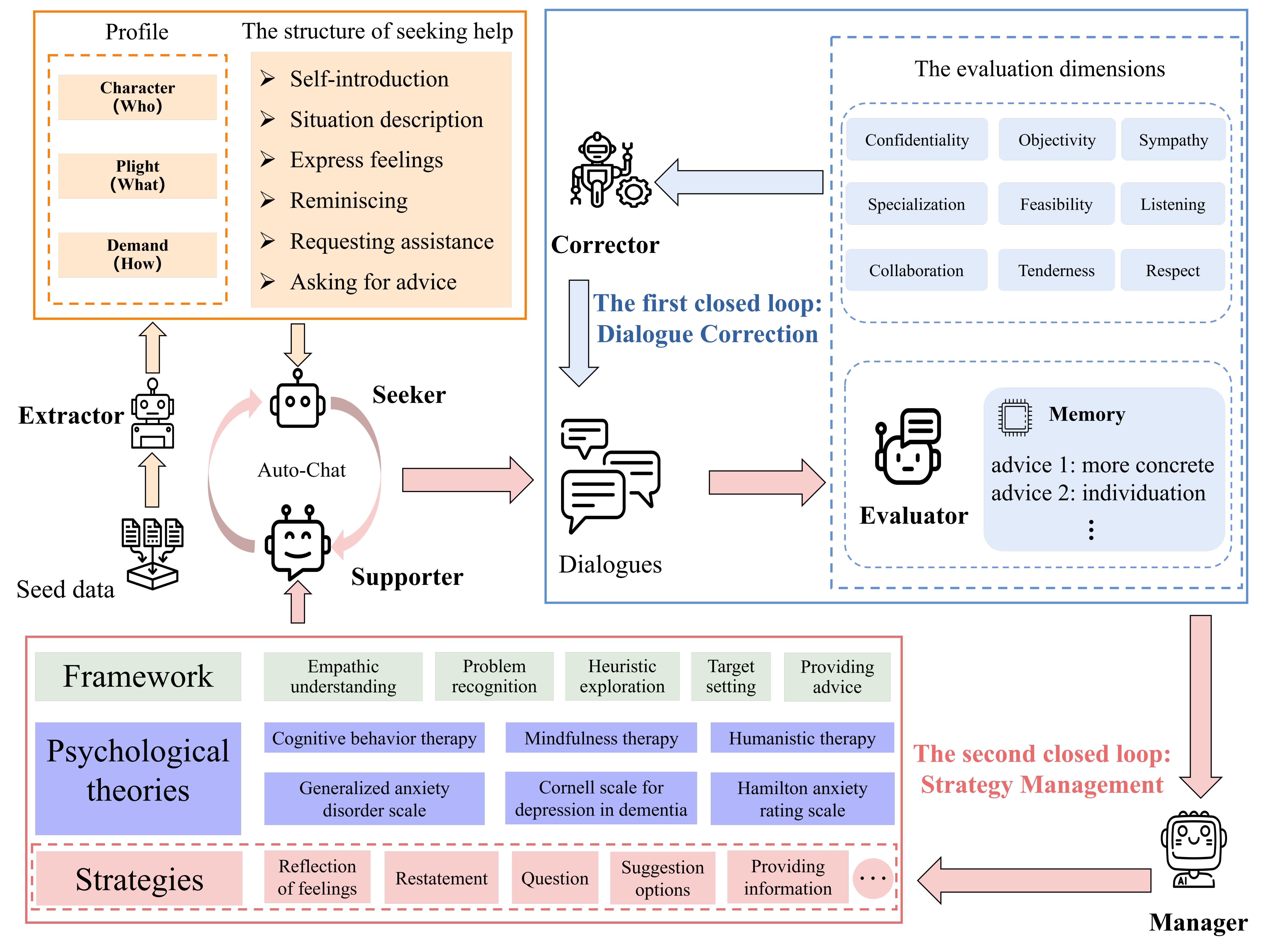}
    \caption{The proposed multi-agent cooperation architecture.}
    \label{proposed_structure}
\end{figure*}

\subsubsection{Dialogue construction process}
We propose a multi-agent architecture comprising six specialized agents: Extractor, Seeker, Supporter, Evaluator, Corrector, and Manager, integrated with a dual-loop feedback mechanism to synthesize emotional support dialogues, as shown in \cref{proposed_structure}. Detailed introductions of these agents are provided in Appendix A, while the corresponding prompt templates used to instantiate each agent are included in Appendix B.

The dialogue generation process is primarily driven by multi-agent collaboration and AI feedback. First, the Extractor retrieves the basic information of the Seeker from a seed session and initializes the Seeker agent. Then, the Supporter and Seeker engage in a multi-turn dialogue simulation. During each turn, the Evaluator assesses the utterance of the Supporter against nine predefined quality dimensions to determine whether revision is needed or the conversation should continue. If revisions are required, the Corrector improves the utterance based on the feedback of the Evaluator. This forms the first feedback loop, ensuring dialogue coherence and rationality within a session. Concurrently, modification suggestions from each consultation session are retained in the memory of the Evaluator. Once the Evaluator determines that the dialogue is complete, the Manager is activated to enrich the psychological support strategies of the Supporter by analyzing the accumulated feedback. This forms the second feedback loop, guaranteeing that the counseling strategies of the Supporter improve incrementally with each consultation.

\subsection{The Training Framework}
\begin{figure*}[!ht]
\centering
    \includegraphics[width=1.6\columnwidth]{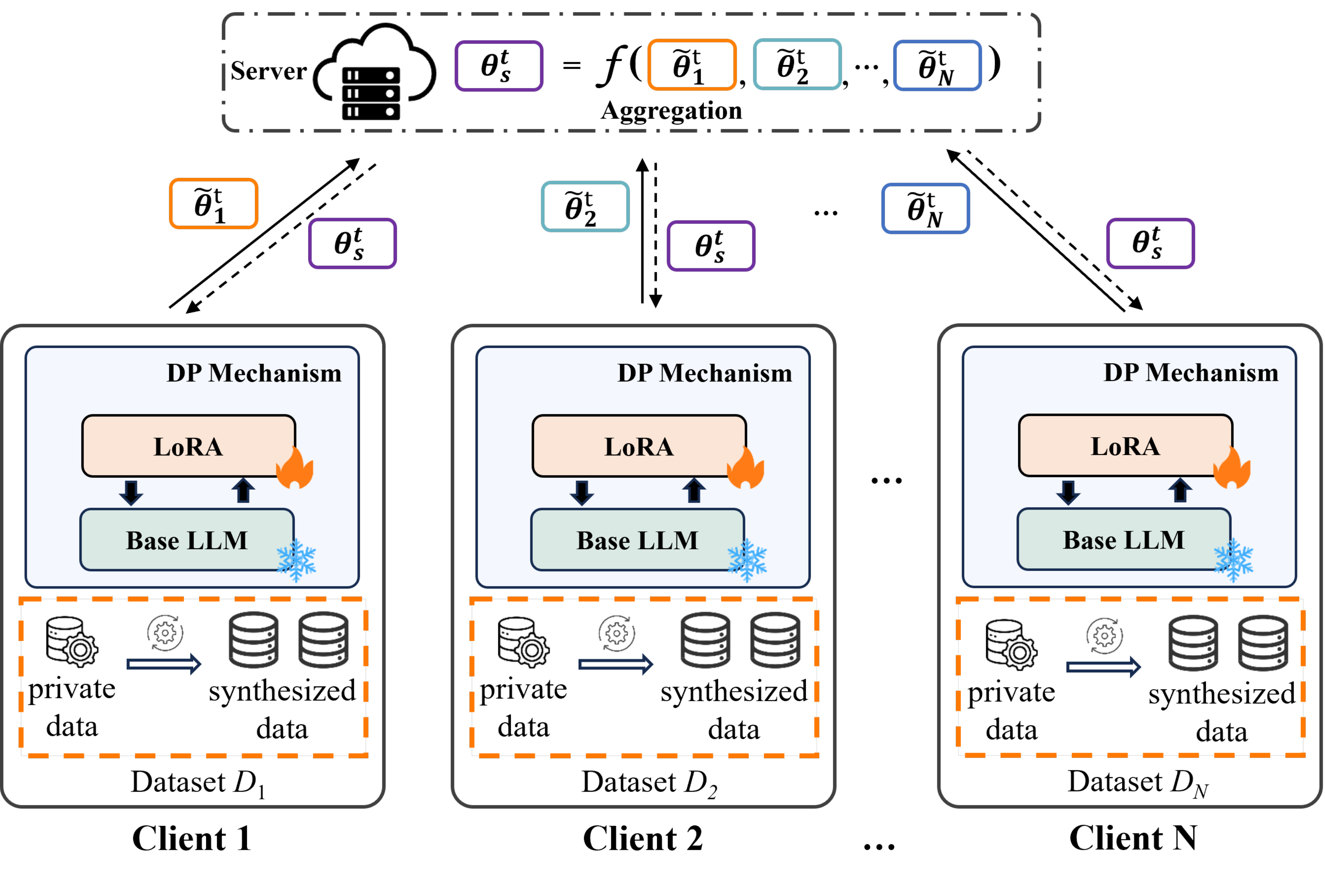}
    \caption{The parameter-efficient privacy-preserving fine-tuning framework.}
    \label{training_framework}
\end{figure*}
The model of \emph{MindChat} is trained within a parameter-efficient and privacy-preserving FL framework, as illustrated in \cref{training_framework}. The design integrates LoRA for efficient on-device fine-tuning and client-level DP to safeguard sensitive mental health dialogues during collaborative training.
\subsubsection{Efficient fine-tuning}
To enable scalable and communication-efficient adaptation of the large base model, \emph{MindChat} adopts the LoRA method. This approach is motivated by empirical observations that weight updates during fine-tuning of LLMs often reside in a low-dimensional subspace. Instead of updating all parameters of the pre-trained model, LoRA freezes the original weights and introduces trainable low-rank decomposition matrices to approximate the update.

Formally, for client $C_i$, the original model parameter matrix $\bm{W}^0 \in \mathbb{R}^{d \times k}$ remains frozen throughout training. At communication round $t$, two trainable matrices $\bm{A}_i^t \in \mathbb{R}^{r \times k}$ and $\bm{B}_i^t \in \mathbb{R}^{d \times r}$ are introduced such that
\begin{equation}
\begin{aligned}
\bm{W}^0 + \Delta \bm{W}_i^t = \bm{W}^0 + \bm{B}_i^t \bm{A}_i^t,
\end{aligned}\label{FL optimize base}
\end{equation}
where $\Delta \bm{W}_i^t = \bm{B}_i^t \bm{A}_i^t$ denotes the LoRA-induced update. Since one can obtain that $r \ll \min(d,k)$, the scale of trainable parameters is significantly reduced, resulting in lower computation and communication costs.

\subsubsection{Privacy protection federated training}
To protect sensitive counseling data during federated training, client-level $(\epsilon,\delta)$-DP is enforced on the model updates transmitted to the server, where $\epsilon$ controls the privacy budget and $\delta$ denotes the probability of privacy failure. Each client $C_i$ holds a frozen base model $\bm{W}^0$ and, at communication round $t$, generates its update by optimizing only the low-rank adapter matrices $\bm{A}_i^t$ and $\bm{B}_i^t$. 
Let $\bm{\theta}_i^t = \{\bm{A}_i^t, \bm{B}_i^t\}$ denote the collection of all trainable parameters in the LoRA adapter matrices on client $C_i$ at round $t$, and let $\bm{\theta}_s^{t-1}$ denote the corresponding global LoRA parameters received from the server. After local training, the client computes its LoRA parameter update as 
$\Delta \bm{\theta}_i^t = \bm{\theta}_i^t - \bm{\theta}_s^{t-1}$.
To control the magnitude of the client update, we apply $L_2$-norm clipping to $\Delta \bm{\theta}_i^t$ with a pre-specified threshold $C$:
\begin{equation}
\text{Clip}_C(\Delta \bm{\theta}_i^t) = \Delta \bm{\theta}_i^t \cdot \min\left(1, \frac{C}{\|\Delta \bm{\theta}_i^t\|_2}\right),
\end{equation}
Gaussian noise is then independently added to each clipped LoRA parameter tensor to ensure the $(\epsilon,\delta)$-DP guarantee. The privatized LoRA update is given by
\begin{equation}
\Delta \tilde{\bm{\theta}}_i^t = \text{Clip}_C(\Delta \bm{\theta}_i^t) + \mathcal{N}(0, \sigma^2 \mathbf{I}),
\end{equation}
where the noise scale $\sigma$ is calibrated according to the Gaussian mechanism as
\begin{equation}
\sigma = \frac{S \sqrt{2 \ln(1.25 / \delta)}}{\epsilon}.
\end{equation}
with $S$ denoting the assumed $L_2$-sensitivity of the client update, which is used to calibrate the Gaussian noise.
The client then sends the perturbed LoRA parameters
$\tilde{\bm{\theta}}_i^t = \bm{\theta}_s^{t-1} + \Delta \tilde{\bm{\theta}}_i^t$
to the server.

Upon receiving the privacy-preserving LoRA parameters from participating clients, the central server aggregates them using the FedAvg algorithm~\cite{mcmahan2017communication}. Specifically, at round $t$, the global LoRA parameters are updated as
\begin{equation}
\bm{\theta}_s^{t}=\sum_{i=1}^{N}\frac{|D_i|}{\sum\limits _{j=1}^{N}|D_j|}\tilde{\bm{\theta}}_i^{t},
\label{fedavg}
\end{equation}
where $|D_i|$ denotes the amount of local training samples on client $C_i$. The aggregated global LoRA parameters $\bm{\theta}_s^{t}$ are then broadcast to all clients to initialize the next round of training.
After the final communication round, the resulting LoRA parameters
$\bm{\theta}_s^{t}=\{\bm{A}_s^{t},\bm{B}_s^{t}\}$ are combined with the frozen base
model to form the complete model weight matrix
$\bm{W}_s^{t} = \bm{W}^0 + \bm{B}_s^{t}\bm{A}_s^{t}$.

\newcolumntype{C}[1]{>{\centering\arraybackslash}p{#1}}

\begin{table*}[t]
\caption{Details of the Five Dimensions in Data Evaluation.}
\label{tab:evaluation_dimensions}
\centering
\renewcommand{\arraystretch}{1.2}
\begin{tabular}{|C{0.10\textwidth}|p{0.85\textwidth}|}
\hline
\multicolumn{2}{|c|}{\textbf{Assessing from the perspective of a supporter}} \\
\hline
\multicolumn{2}{|l|}{\textbf{Professionalism}: flexibility in applying techniques and dynamic adjustment of strategies} \\
\hline
\multirow{2}{=}[-2.8ex]{\centering Details} 
& whether the supporter flexibly applies psychological counseling techniques based on established theories to help the seeker identify and address their problems. \\
\cline{2-2} 
& whether the supporter is able to dynamically adjust its counseling strategies and techniques in response to the emotional state and evolving needs of the seeker. \\
\hline
\multicolumn{2}{|l|}{\textbf{Helpfulness}: effective emotional support} \\
\hline
\multirow{2}{=}[-1.8ex]{\centering Details} 
& whether the supporter can effectively relieve the client's emotional pressure and help him or her clarify the direction or steps of problem solving. \\
\cline{2-2} 
& whether the supporter can provide practical and effective support and avoids formal comfort. \\
\hline
\multicolumn{2}{|l|}{\textbf{Guidance}: clear and realistic goals} \\
\hline
\multirow{2}{=}[-1.8ex]{\centering Details} 
& whether the supporter offers clear suggestions and feasible goals. \\
\cline{2-2} 
& whether the supporter ensures the recommendations are practical and can be implemented by the seeker leading to tangible improvements in their life. \\
\hline
\multicolumn{2}{|c|}{\textbf{Assessing from the perspective of a seeker}} \\
\hline
\multicolumn{2}{|l|}{\textbf{Emotion}: perception and adjustment} \\
\hline
\multirow{2}{=}[-1.8ex]{\centering Details} 
& whether the seeker shows normal emotional fluctuations during the communication process, avoiding emotional monotony or dullness. \\
\cline{2-2} 
& whether the seeker controls emotional responses and avoids being overly negative or pessimistic. \\
\hline
\multicolumn{2}{|l|}{\textbf{Trust}: building trust and a sense of security} \\
\hline
\multirow{2}{=}[-1.8ex]{\centering Details} 
& whether the seeker gradually builds a trusting relationship with the supporter through transparent, honest, and friendly communication. \\
\cline{2-2} 
& whether the seeker feels accepted and is willing to express true thoughts during the communication. \\
\hline
\end{tabular}
\end{table*}

\subsection{Evaluation Methods}
To evaluate the generated data and models, a combination of automatic and human assessments is adopted. For automatic evaluation, the GPT-4o model is employed to score the generated data and trained models based on predefined criteria, with a fixed evaluation prompt and consistent decoding settings, including temperature, across all evaluations. Additionally, human evaluation is conducted by four postgraduate students with expertise in psychology. 

The quality of conversational corpora directly influences the performance of models trained on it. In research on evaluating synthetic data for mental health assistance, current evaluation dimensions are typically supporter-centered, focusing on reasonable response of the supporter~\cite{zheng-etal-2024-self}. However, in reality, effectiveness of psychological consultations is ultimately determined by feelings of the seeker. Inspired by~\cite{xiao-etal-2024-healme, Hill_helping_20}, this paper proposes five evaluation dimensions, Professionalism~(Pro.), Helpfulness~(Hel.), Guidance~(Gui.), Emotion~(Emo.), and Trust~(Tru.), to comprehensively assess the generated data from both seeker and supporter perspectives. These dimensions are thoroughly explained in \cref{tab:evaluation_dimensions}.

Particularly, CpsyCounE~\cite{zhang-etal-2024-cpsycoun}, a dataset designed to evaluate the psychological counseling capabilities of LLMs, is utilized to assess \emph{MindChat} and baselines. According to~\cite{zhang-etal-2024-cpsycoun}, the evaluation is performed from four dimensions: \textit{Comprehensiveness}, \textit{Professionalism}, \textit{Authenticity}, and \textit{Safety}. \textit{Comprehensiveness} refers to the extent to which the dialogue covers the background of users and psychological concerns, \textit{Professionalism} indicates the proficiency of the model in psychological counseling, \textit{Authenticity} reflects the degree to which the dialogue aligns with real-world scenarios and \textit{Safety} denotes the level of privacy protection provided to the user. In addition, we adopt ROUGE-1 and cosine similarity as proxies for data memorization, and further assess privacy guarantees through membership inference attacks (MIAs), quantified using both ROC AUC~\cite{murakonda2021quantifying} and PR AUC~\cite{shokri2017membership}. These metrics collectively reflect the strength of privacy protection under DP.

\section{Experiments}\label{experiment}
In this section, a systematic experimental evaluation of the proposed \emph{MindChat} framework is presented. The experimental setup and training configurations are first introduced. Subsequently, the quality of the synthesized \emph{MindCorpus} and the effectiveness of the proposed dual-loop multi-agent data construction framework are analyzed, including comparisons of coordination strategies among agents employing different LLMs and ablation studies on agent collaboration. Comparative evaluations with both general-purpose and counseling-oriented LLMs are then conducted. Finally, the impact of data scale, topical diversity, and privacy-preserving mechanisms on model performance is investigated.

\subsection{Baselines}
For the dataset-level comparison, we evaluate \emph{MindCorpus} against seven existing emotional dialogue datasets. These include four Chinese datasets: PsyDTCorpus~\cite{xie2025psydt}, SoulChatCorpus~\cite{chen-etal-2023-soulchat}, SMILECHAT~\cite{qiu-etal-2024-smile}, and CPsyCounD~\cite{zhang-etal-2024-cpsycoun}, as well as three English datasets: ExTES~\cite{zheng-etal-2024-self}, ESD-CoT~\cite{zhang-etal-2024-escot}, and AUGESC~\cite{zheng-etal-2023-augesc}.

At the model level, \emph{MindChat} is benchmarked against a diverse set of LLMs. This set includes widely used general-purpose models such as ChatGPT~\cite{achiam2023gpt}, DeepSeek~\cite{liu-2024-deepseek}, and Gemini~\cite{team-2024-gemini}, the base model Qwen3-8B~\cite{yang2025qwen3}, and four specialized mental health LLMs SoulChat2.0~\cite{chen-etal-2023-soulchat}, CPsyCounX~\cite{zhang-etal-2024-cpsycoun}, EmoLLM2.0~\cite{yang2024emollm}, and MeChat~\cite{qiu-etal-2024-smile}.

\subsection{Experimental Settings}
\begin{figure}[!ht]
    \centering
    \includegraphics[width=1.0\columnwidth]{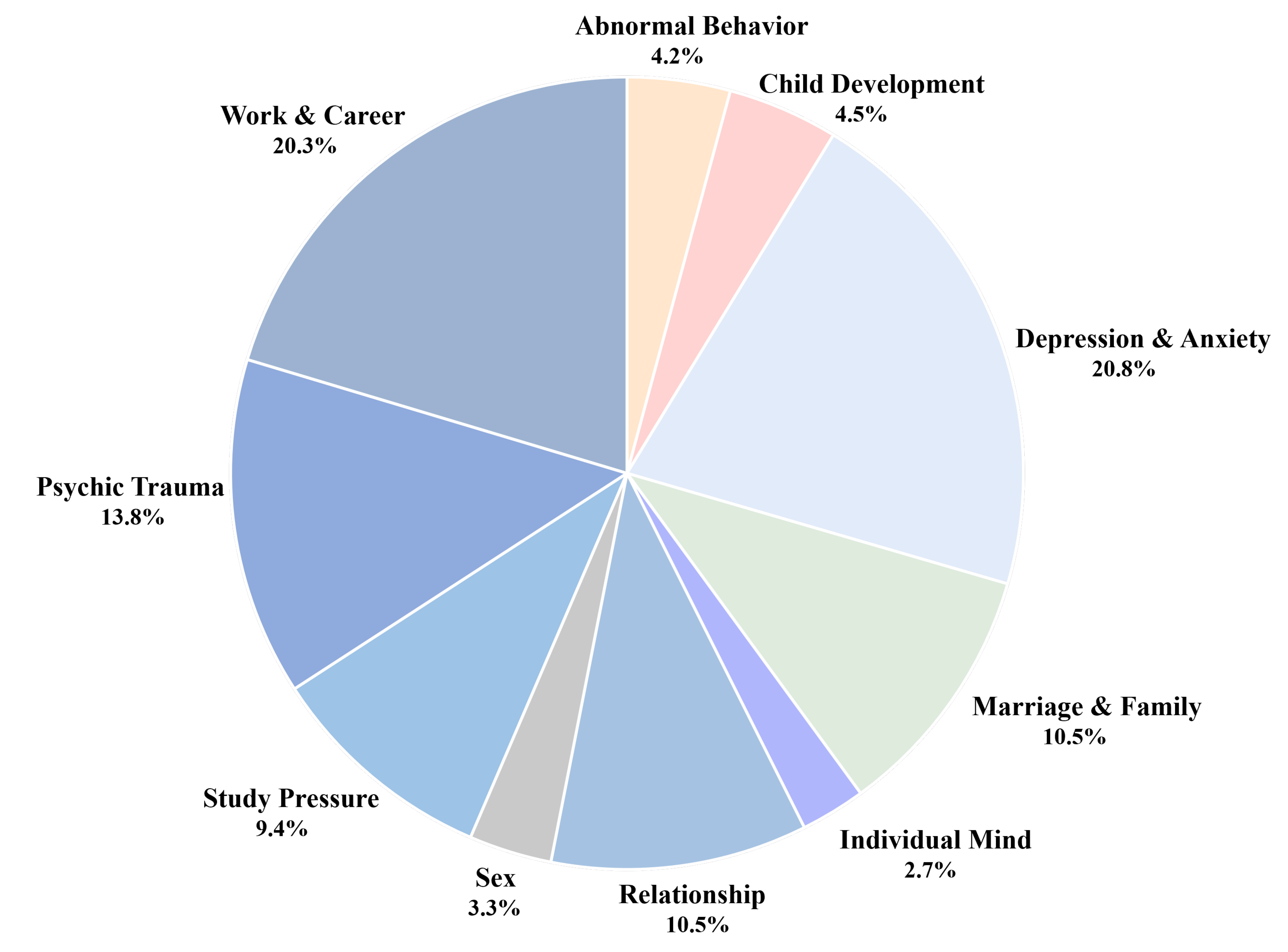}
    \caption{Data distribution of \emph{MindCorpus}.}
    \label{fig:enter-label}
\end{figure}
\emph{MindChat} is built upon the Qwen3-8B base model and trained on a single NVIDIA A800 GPU using \emph{MindCorpus}. The base training configuration includes a batch size of $16$, LoRA rank $16$, LoRA alpha $32$, and a maximum sequence length of $512$. The learning rate is kept constant and selected from $[1e^{-6}, 5e^{-5}]$ based on validation performance. In the federated training process, we set the number of clients to $10$, corresponding to the thematic categories in \emph{MindCorpus}. The training runs for a total $100$ communication rounds, and in each round, every participating client performs $3$ epochs of local training on its private data. We apply $4$-bit quantization to reduce memory overhead. Local differential privacy is implemented at the client level by perturbing the model update before communication. Each client clips its local LoRA-based model update to a maximum $L_2$ norm of $1$ and applies Gaussian noise with sensitivity set to $1$, yielding an $(\epsilon,\delta)$-DP guarantee with $\epsilon = 1$ and $\delta = 10^{-5}$. Unless otherwise specified, all experiments follow the above configurations.

\subsection{Results}
\subsubsection{Synthesized Data Analysis}
The synthesized dataset \emph{MindCorpus} contains $5.7$k dialogue sessions across diverse themes, generated from curated scenario scripts using a multi-agent collaborative framework. As shown in \cref{fig:enter-label}, dialogues are distributed across multiple themes. Table \ref{table:basic_stats} summarizes the basic statistics of \emph{MindCorpus} in comparison with existing emotional dialogue datasets. 

\begin{table*}[t]
\setlength{\tabcolsep}{34pt}
\caption{Basic Statistics of Emotional Dialogue Datasets.}
\label{table:basic_stats}
\centering
\begin{tabularx}{\linewidth}{@{}ccccc@{}}
\toprule
\textbf{Dataset} & \textbf{Language} & \textbf{Sessions} & \textbf{Utterances} & \textbf{Length} \\
\midrule
PsyDTCorpus    & Chinese & 5k    & 18.1 & 89.7 \\
SoulChatCorpus & Chinese & 258k  & 5.9  & 131.4 \\
SMILECHAT      & Chinese & 56k   & 10.4 & 55.0 \\
CPsyCounD      & Chinese & 3.1k  & 8.0  & 85.5 \\
ExTES          & English & 11.2k & 18.2 & 26.0 \\
ESD-CoT        & English & 1.7k  & 23.4 & 18.5 \\
AUGESC         & English & 65k   & 26.7 & 37.2 \\
MindCorpus     & Chinese & 5.7k  & 12.0 & 84.0 \\
\bottomrule
\end{tabularx}
\end{table*}

\begin{table*}[t]
\setlength{\tabcolsep}{12pt}
\caption{Automatic and Human Evaluations of Emotional Dialogue Datasets. Highest Scores are in \textbf{bold}.}
\label{table:evaluation_scores}
\begin{tabularx}{\linewidth}{@{}ccccccccccc@{}}
\toprule
\multirow{2}{*}{\textbf{Dataset}} & \multicolumn{5}{c}{\textbf{Automatic Evaluation}} & \multicolumn{5}{c}{\textbf{Human Evaluation}} \\
\cmidrule(r){2-6} \cmidrule(l){7-11}
                 & Pro. & Hel. & Gui. & Emo. & Tru. & Pro. & Hel. & Gui. & Emo. & Tru. \\
\midrule
PsyDTCorpus      & 8.92 & 8.90 & 8.84 & 8.86 & 8.28 & 8.41 & 8.65 & 8.15 & \textbf{9.12} & 8.43 \\
SoulChatCorpus   & 8.09 & 8.16 & 8.14 & 8.11 & 7.78 & 8.41 & 8.43 & 7.95 & 8.77 & 7.97 \\
SMILECHAT        & 8.14 & 8.23 & 8.10 & 8.36 & 7.86 & 8.11 & 8.15 & 8.26 & 8.57 & 7.60 \\
CPsyCounD        & 8.60 & 8.58 & 8.62 & 8.65 & 8.03 & 8.17 & 8.21 & 8.13 & 8.93 & 8.33 \\
ExTES            & 8.66 & 8.76 & 8.68 & 8.86 & \textbf{8.38} & 8.12 & 8.36 & 8.34 & 8.87 & 8.21 \\
ESD-CoT          & 8.40 & 8.38 & 8.26 & 8.58 & 8.14 & 7.44 & 7.49 & 7.21 & 8.63 & 7.43 \\
AUGESC           & 5.12 & 4.98 & 4.70 & 6.64 & 6.08 & 6.31 & 6.85 & 6.57 & 7.27 & 6.77 \\
MindCorpus       & \textbf{8.94} & \textbf{8.98} & \textbf{8.96} & \textbf{8.98} & 8.32 & \textbf{8.45} & \textbf{8.68} & \textbf{8.35} & 8.93 & \textbf{8.50} \\
\bottomrule
\end{tabularx}
\end{table*}

\begin{table*}[!ht]
\setlength{\tabcolsep}{24pt}
\caption{Spearman Rank Correlations between Automatic and Human Evaluations across Emotional Dialogue Datasets, Reported per Evaluation Dimension.}
\label{table:Spearman_data}
\begin{tabularx}{\linewidth}{@{}cccccc@{}}
\toprule
\textbf{Spearman correlation statistics} & \textbf{Pro.} & \textbf{Hel.} & \textbf{Gui.} & \textbf{Emo.} & \textbf{Tru.} \\
\midrule
$\rho$      & 0.659          & 0.738          & 0.690          & 0.819          &0.667          \\
$p$-value   & 0.076          & 0.037          & 0.058          & 0.013          &0.071          \\
\bottomrule
\end{tabularx}

    \smallskip
    {\raggedright
    Correlation strength: $\rho \in [0.30, 0.49]$: low, $\rho \in [0.50, 0.69]$: moderate, $\rho \in [0.70, 0.89]$: high. Correlations with $p$-value $< 0.10$ are considered statistically significant.\par}
\end{table*}

\begin{table*}[!ht]
    \setlength{\tabcolsep}{1.0pt}
    \caption{Evaluation Results for Data Synthesized by Heterogeneous Multi-agent Groups. Highest Scores are in \textbf{bold}, Second-highest are \underline{underlined}.} 
    \label{tab:multi-agent}
    \begin{tabular*}{1.0\textwidth}{@{\extracolsep{\fill}}cccccccccccccc}
    \toprule
        Gro. & See. & Sup. & Eva. & Cor. & Man. & Mod.(\%) & Spe. & Len. & Pro. & Hel. & Gui. & Emo. & Tru. \\
        \midrule
        1 & 0.5B & 0.5B & 7B & 3B & 3B & 22.39 & 6.76 & 316.84 & 6.20 & 6.13 & 6.00 & 6.80 & 6.73 \\
        2 & 1.5B & 1.5B & 7B & 3B & 3B & 4.26 & 3.86 & 161.45 & 8.60 & 8.67 & 8.73 & 8.73 & 8.07 \\
        3 & 1.5B & 1.5B & 7B & 7B & 3B & 2.00 & 6.40 & 191.20 & 8.73 & 8.73 & 8.73 & 8.87 & 8.13 \\
        4 & 1.8B & 1.8B & 7B & 3B & 3B & 7.84 & 9.83 & 558.67 & 5.27 & 5.00 & 5.00 & 5.33 & 4.87 \\
        5 & 4B & 4B & 7B & 3B & 3B & 2.82 & 8.09 & 189.01 & \underline{8.87} & \underline{8.93} & \underline{8.93} & \textbf{9.00} & 8.13 \\
        6 & 3B & 3B & 7B & 3B & 3B & 13.64 & 2.67 & 58.03 & 8.00 & 8.07 & 8.33 & 8.47 & 8.00 \\
        7 & 7B & 7B & 7B & 3B & 3B & 8.60 & 2.32 & 57.93 & 8.40 & 8.53 & 8.67 & 8.73 & \underline{8.20} \\
        8 & 7B & 7B & 7B & 7B & 3B & 8.77 & 3.02 & 54.93 & 8.80 & 8.80 & 8.87 & 8.93 & \underline{8.20} \\
        9 & 14B & 14B & 7B & 3B & 3B & 4.55 & 4.09 & 55.76 & \underline{8.87} & 8.87 & \underline{8.93} & \textbf{9.00} & 8.00 \\
        10 & 72B & deepseek & gpt & glm & qwen & 18.9 & 6.22 & 69.02 & 8.53 & 8.60 & 8.53 & 8.86 & 7.78 \\
        11 & 72B & qwen & gpt & glm & qwen & 7.05 & 13.73 & 84.00 & \textbf{8.94} & \textbf{8.98}& \textbf{8.96} & \underline{8.98} & \textbf{8.32} \\
        \bottomrule                 
    \end{tabular*}

    \smallskip
    {\raggedright
    Gro., See., Sup., Eva., Cor., and Man. are abbreviations for Group, Seeker, Supporter, Evaluator, Corrector, and Manager, respectively. Mod. denotes the modification ratio of utterances during each session. Spe. indicates the average generation time per session in seconds. Len. represents the average length per utterance. 1.8B and 4B refer to InternLM2.5-1.8B-Chat and MiniCPM3-4B, respectively. All other size-labeled models belong to the Qwen2.5 Instruct series, for instance, 1.5B refers to Qwen2.5-1.5B-Instruct. Additionally, deepseek refers to the DeepSeek-V3 model, glm represents the GLM-4-Plus model, gpt stands for the GPT-4o model, and qwen denotes the Qwen-Max model.\par}
\end{table*}

To evaluate data quality, we follow the sampling protocol of~\cite{zheng-etal-2024-self}, randomly selecting $50$ sessions from each dataset while ensuring diversity by incorporating data from clearly classified content categories. We employ GPT-4o as a judge model to score dialogues based on our five-dimensional evaluation framework. Table \ref{table:evaluation_scores} reports the results of both automated and human evaluations. The results of automatic evaluation show that \emph{MindCorpus} attain the best performance on all metrics except for Tru. In human evaluation, aside from Emo., the remaining four metrics consistently indicate that \emph{MindCorpus} outperforms other emotional dialogue datasets. Overall, both automated and human evaluations confirm the superior quality of \emph{MindCorpus} compared to existing emotional dialogue datasets. To further examine the consistency between automatic and human evaluations, Spearman rank correlations~\cite{zar2005spearman} are computed for each evaluation dimension at the dataset level. As shown in Table~\ref{table:Spearman_data}, most dimensions exhibit moderate to high agreement, demonstrating strong consistency between automated and human evaluations and supporting the reliability of the reported results in Table~\ref{table:evaluation_scores}.

\subsubsection{Heterogeneous Multi-LLM Coordination}
We examine the performance of heterogeneous LLMs collaborating within the proposed multi-agent framework, including models of varying sizes and different developers. Table~\ref{tab:multi-agent} presents the performance of these LLMs working together in $11$ coordination groups. 

During the experiments, we observe that excessively long model outputs tend to exhibit content homogenization, which deviates from real-world psychological counseling scenarios. Empirical statistics of publicly available emotional and counseling dialogue datasets in Table~\ref{table:basic_stats} indicate that the average length of a single response is generally below $100$ words. Motivated by this distribution, we treat dialogues with single-round response lengths under $100$ words as qualified data. Accordingly, Groups $6$ to $11$ in Table~\ref{tab:multi-agent} meet this criterion. 

With different collaboration groups, the proportion of dialogues requiring modification changes, implying various interactions between agents. More interactions typically result in longer durations. As shown in Groups 5 and 6, compared to Qwen2.5-3B-Instruct, MiniCPM-4B produces fewer qualified dialogues and generates relatively longer responses. However, GPT-4o-based evaluation scores are higher for longer responses, possibly because the automated judges evaluate responses using point-based criteria, whereby longer outputs are more likely to satisfy multiple scoring aspects. In general, larger model sizes correlate with higher data quality, though this trend is not strictly monotonic. In Groups 10 and 11, we employ commercial models for data synthesis via API calls. The synthesized data quality in Group 11 surpasses that of all other experimental groups. However, Group 10 underperforms relative to the small-model collaborations in Groups 8 and 9. This discrepancy may be attributed to performance differences between DeepSeek-V3 and Qwen-Max. As shown in Table~\ref{tab:multi-agent}, the quality of data generated through multi-agent collaboration depends not only on model scale but also strongly on model type. Notably, relatively smaller-scale models also demonstrate the potential to achieve performance comparable to that of commercial-grade models.

\begin{figure}[ht]
    \centering
    \includegraphics[width=1.0\columnwidth]{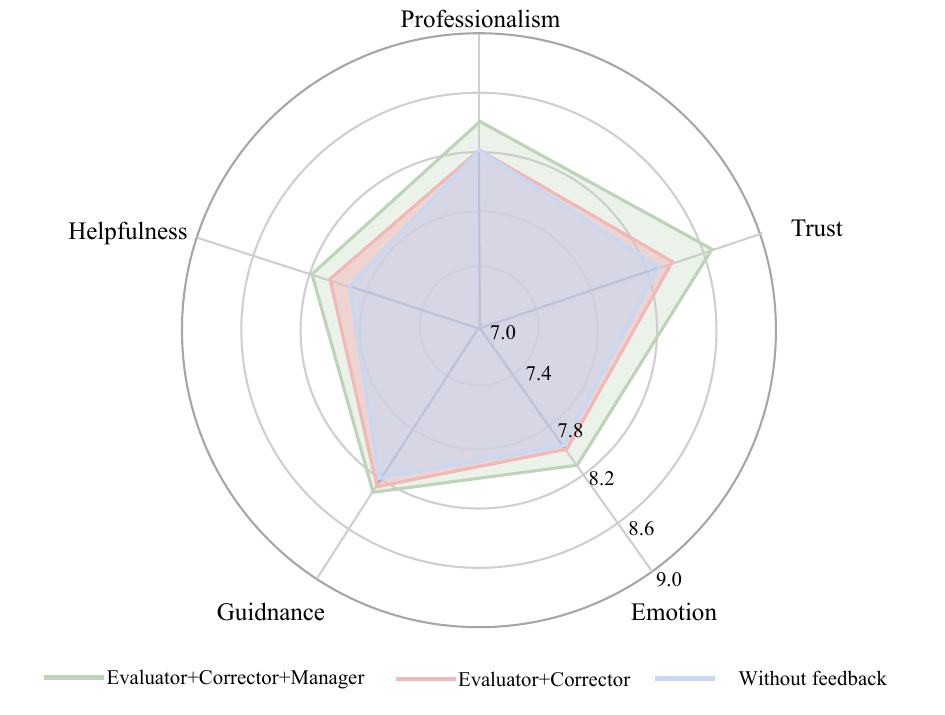}
    \caption{The impact of multi-agent collaboration on the quality of synthetic data.}
    \label{ablation}
\end{figure}

\subsubsection{Impact of Key Agents on Data Quality}
To examine the functions of key agents within the proposed multi-agent framework, we carry out ablation tests targeting the functionalities of the Evaluator, Corrector, and Manager agents. \cref{ablation} displays the data quality assessment results under different agent collaboration configurations. Dialogues generated solely through interactions between Seeker and Supporter, without any form of feedback mechanism, exhibit the lowest quality. The inclusion of the Corrector, which revises inappropriate responses based on Evaluator feedback, leads to a modest improvement in data quality. Additional gains are observed by introducing Manager, which enriches support strategies for Supporter, thereby not only enhancing the professionalism of Supporter but also fostering a stronger sense of trust between Seeker and Supporter.

\begin{figure}[!ht]
    \centering
    \includegraphics[width=1.0\columnwidth]{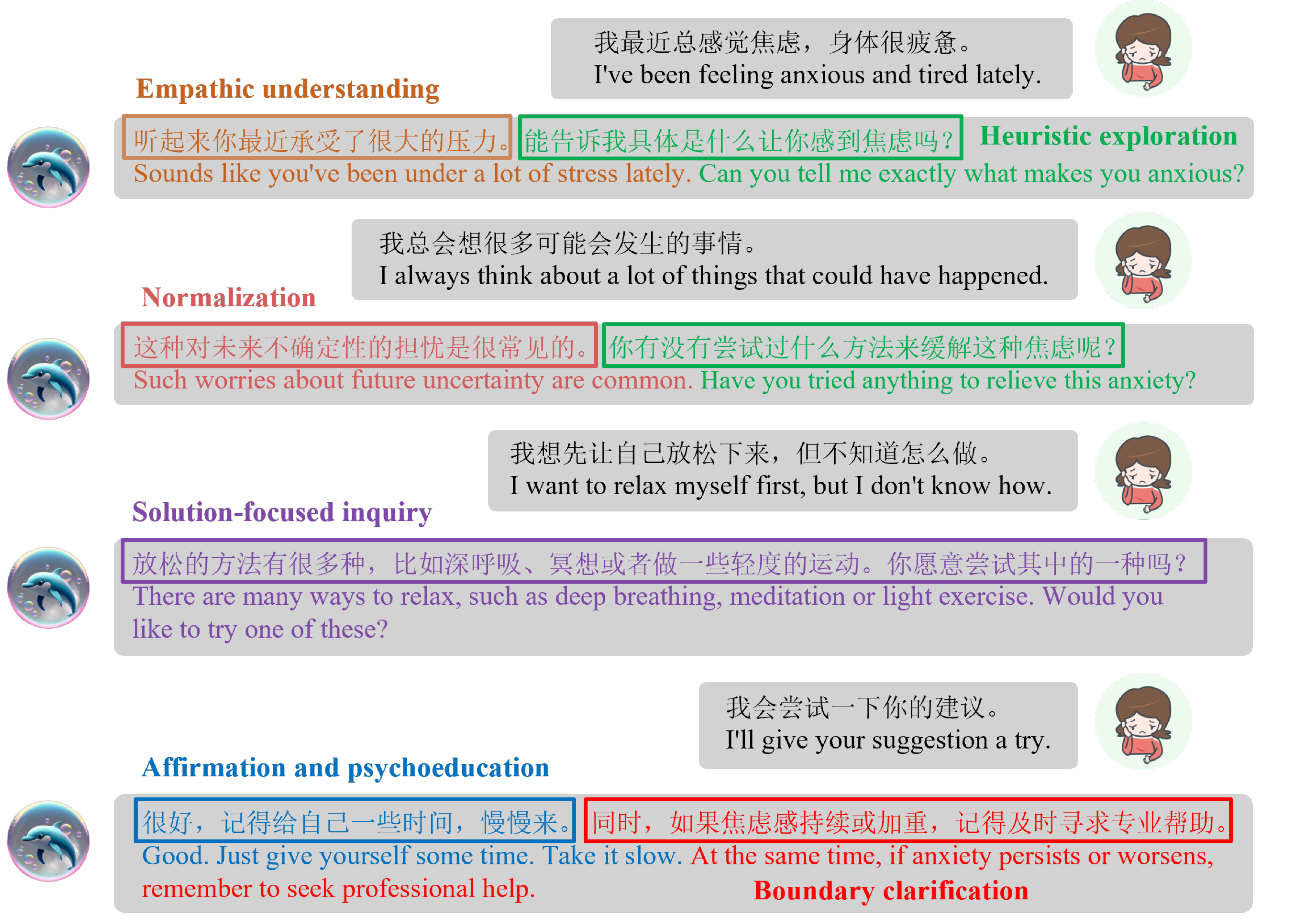}
    \caption{The responses of \emph{MindChat} to inquiries from an anxiety seeker.}
    \label{Case}
\end{figure}
\subsubsection{Comparison of Different Psychological LLMs}
Based on the Qwen3-8B model, we develop \emph{MindChat} by using the artificial emotional multi-round dialogue dataset \emph{MindCorpus} within the proposed privacy-preserving fine-tuning architecture. \cref{Case} presents a representative session generated by \emph{MindChat}, demonstrating its capability to deliver supportive interactions while safeguarding user privacy.

\begin{table*}[!ht]
    \setlength{\tabcolsep}{12.0pt}
    \caption{Automatic and Human Evaluations of Multiple LLMs in Psychological Counseling Capabilities.}
    \label{table:evaluation_model}
    \begin{tabularx}{\textwidth}{@{} c *{10}{c} @{}}
        \toprule
        \multirow{2}{*}{\textbf{Model}} & \multicolumn{5}{c}{\textbf{Automatic Evaluation}} & \multicolumn{5}{c}{\textbf{Human Evaluation}}\\
        \cmidrule(lr){2-6} \cmidrule(lr){7-11}
         & Com. & Pro. & Aut. & Saf. & Avg. & Com. & Pro. & Aut. & Saf. & Avg. \\ 
        \midrule
        ChatGPT &  1.59 & 2.40 & 2.46 & 1.00 & 1.86 & 1.36 & \underline{2.26} & 2.18 & 0.90 & 1.67 \\
        DeepSeek &  \underline{1.74} & 2.61 & 2.61 & 1.00 & 1.99 & \textbf{1.80} & \textbf{2.40} & \textbf{2.50} & \textbf{0.95} & \textbf{1.91} \\
        Gemini &  1.72 & \underline{2.62} & \underline{2.67} & 1.00 & \underline{2.00} & \textbf{1.80} & 2.00 & 2.30 & \textbf{0.95} & \underline{1.76} \\
        Qwen3-8B & 1.72 & 2.58 & 2.59 & 1.00 & 1.97 & 1.60 & 1.65 & 1.02 & 0.80 & 1.27 \\ 
        SoulChat2.0 &  1.26 & 2.06 & 2.47 & 1.00 & 1.70 & 1.53 & 1.70 & 1.52 & 0.70 & 1.36 \\
        CPsyCounX &  1.64 & 2.35 & 2.42 & 1.00 & 1.85 & 1.27 & 1.63 & 1.38 & 0.82 & 1.28 \\
        EmoLLM2.0 &  1.58 & 2.45 & \underline{2.67} & 1.00 & 1.93 & 0.70 & 1.06 & 1.08 & 0.92 & 0.94 \\ 
        MeChat &  1.64 & 2.34 & 2.59 & 1.00 & 1.90 & 1.40 & 2.00 & 1.82 & 0.80 & 1.50 \\ 
        MindChat &  \textbf{1.91} & \textbf{2.85} & \textbf{2.86} & 1.00 & \textbf{2.16} & \underline{1.63} & 2.22 & \underline{2.43} & \underline{0.93} & 1.75 \\
        \bottomrule
    \end{tabularx}
    
    \smallskip
    {\raggedright
    \emph{Comprehensiveness} ranges from 0 to 2. \emph{Professionalism} varies between 0 and 3. \emph{Authenticity} ranges from 0 to 3. \emph{Safety} varies between 0 and 1.\par}
\end{table*}

In order to evaluate the performance differences between \emph{MindChat} and other benchmark models, we conduct both automatic and human evaluations. The results are summarized in Table~\ref{table:evaluation_model}. In the automatic evaluation, \emph{MindChat} achieves the highest scores across all metrics, better than the general LLMs and the existing mental health-focused LLMs. It is worth noting that \emph{MindChat} demonstrates superior performance in the dimensions of \textit{Comprehensiveness}, \textit{Professionalism}, and \textit{Authenticity}. These three metrics reflect the depth and quality of psychological support provided, thus highlighting the enhanced capability of \emph{MindChat} in addressing the psychological background of seeker, offering professional guidance, and establishing authentic empathetic connections. In the human evaluation, DeepSeek and Gemini, as general LLMs with significantly larger parameter sizes, occupy the top two respectively. Despite this disadvantage in terms of model scale, \emph{MindChat} still ranks third overall. More importantly, it significantly outperforms other models of comparable size, including both general-purpose and domain-specific psychological LLMs, underscoring its effectiveness and competitiveness in mental health scenarios while preserving user privacy.

To further assess the consistency between automatic and human evaluations at the model level, Spearman rank correlations are computed for each evaluation dimension. The results show that \textit{Comprehensiveness} achieves a high correlation of $0.730$ with a p-value of $0.026$, while \textit{Professionalism} and \textit{Authenticity} exhibit lower correlations of $0.351$ with a p-value of $0.354$ and $0.378$ with a p-value of $0.316$, respectively. The observed correlations show clear consistency between automated and human evaluations in \textit{Comprehensiveness}, whereas \textit{Professionalism} and \textit{Authenticity} exhibit relatively weaker alignment. By offering complementary yet distinct perspectives, the two evaluation schemes jointly strengthen the credibility of the model comparison results reported in Table~\ref{table:evaluation_model}.

\subsubsection{Effects of Data Quantity and Diversity}
\begin{figure*}[!ht]
    \centering
    \subfloat[Quantity slices under same theme.\label{subfig:quantity}]{%
        \includegraphics[width=0.40\textwidth]{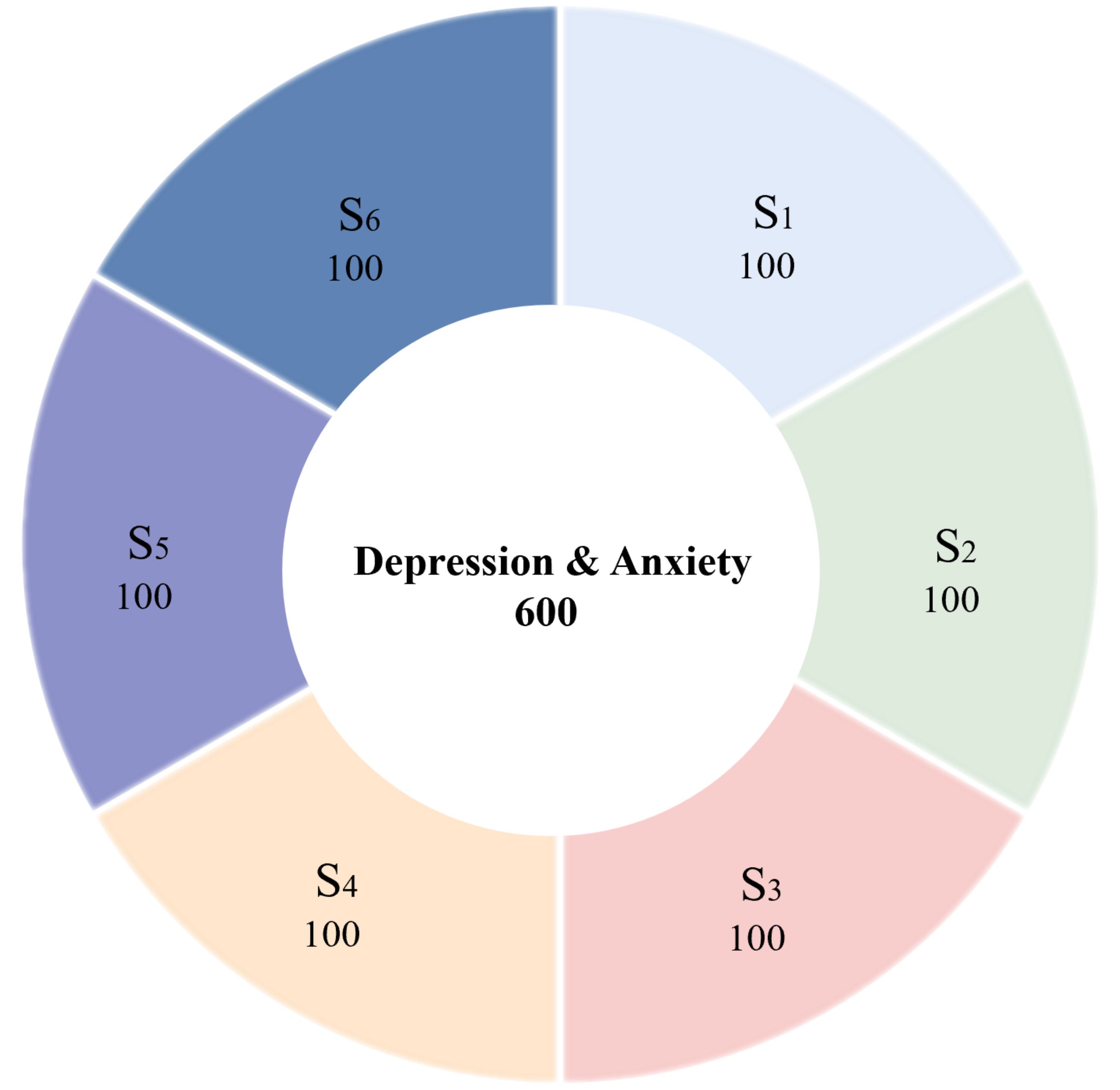}%
    }
    \hfill
    \subfloat[Theme slices with same quantity.\label{subfig:diversity}]{%
        \includegraphics[width=0.40\textwidth]{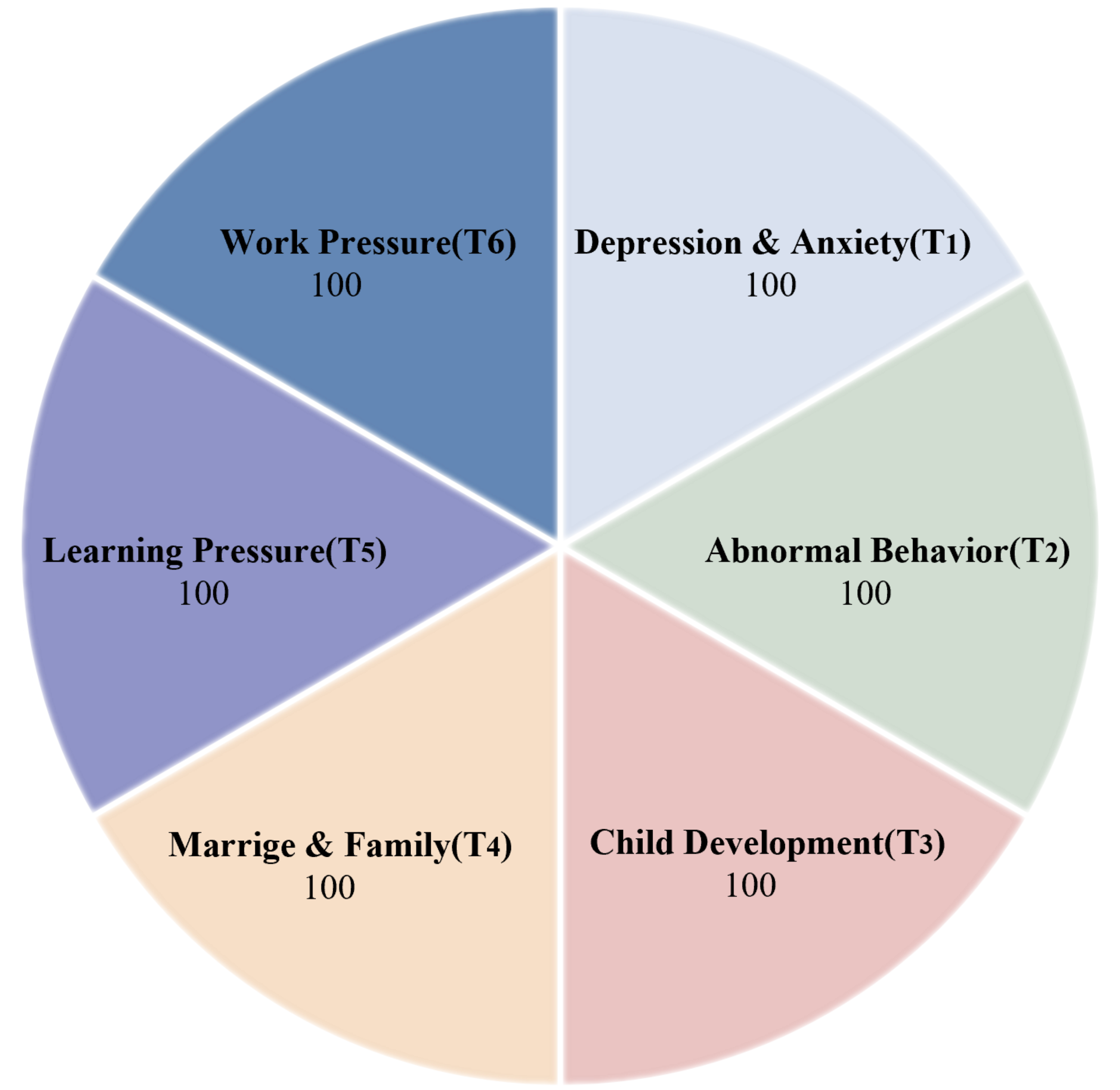}%
    }
    \caption{Comparison of slices: (a) Same theme, (b) Different themes.}
    \label{subfig:slice}
\end{figure*}

In FL architecture, an increasing number of participating clients often leads to larger data quantities and potentially greater diversity. This section investigates how the scale and diversity of training data affect the performance of \emph{MindChat} under the designed training framework. To evaluate the impact of data quantity, we sample 600 sessions from the \emph{Depression and Anxiety} dataset and partition them into six equal subsets, as shown in \cref{subfig:quantity}. For data diversity analysis, we collect 100 samples from each of six distinct themes, outlined in \cref{subfig:diversity}. All experiments are conducted with $\epsilon=1$, and model performance is compared across settings with $2$, $4$, and $6$ collaborating clients. A local baseline is trained using $100$ samples from a single theme.

\begin{table*}[!ht]
    \setlength{\tabcolsep}{20pt}
    \caption{Impact of Data Quantity and Diversity on the Performance of \emph{MindChat}.}
    \label{Comparison of client quantities under the same theme}
    \begin{tabularx}{1.0\textwidth}{@{}cccccccc@{}}
        \toprule
        \multirow{2}{*}{\textbf{Model}} & \multirow{2}{*}{\textbf{N-Clients}} & \multirow{2}{*}{\textbf{Slices}} & \multicolumn{5}{c}{\textbf{Evaluation}}\\
        \cmidrule{4-8}
         & & & Com. & Pro. & Aut. & Saf. & Avg.\\
        \midrule
        MindChat-Local & 1 & ${S_1(T_1)}$ & 1.76 & 2.55 & 2.64 & 1.00 & 1.99 \\        
        \specialrule{0.5pt}{0.1pt}{0pt}
        \multicolumn{7}{c}{\textit{\small Comparison of Client Quantity  }} \\
        \midrule
        MindChat-Q2 & 2 & ${S_1 \sim S_2}$ & 1.73 & 2.57 & 2.60 & 1.00 & 1.98 \\
        MindChat-Q4  & 4 & ${S_1 \sim S_4}$ & \textbf{1.82} & 2.66 & 2.73 & 1.00 & 2.05\\
        MindChat-Q6  & 6 & ${S_1 \sim S_6}$ & \textbf{1.82} & \textbf{2.68} & \textbf{2.75} & 1.00 & \textbf{2.06} \\
        \specialrule{0.5pt}{0.1pt}{0pt}
        \multicolumn{7}{c}{\textit{\small Comparison of Client Theme}} \\
        \midrule
        MindChat-T2 & 2 & ${T_1 \sim T_2}$ & 1.84 & 2.67 & 2.76 & 1.00 & 2.07 \\
        MindChat-T4  & 4 & ${T_1 \sim T_4}$ & 1.86 & 2.69 & 2.80 & 1.00 & 2.09 \\
        MindChat-T6  & 6 & ${T_1 \sim T_6}$ & \textbf{1.89} & \textbf{2.77} & \textbf{2.85} & 1.00 & \textbf{2.13}  \\
        \bottomrule         
    \end{tabularx}

    \smallskip
    {\raggedright
    $Q$ and $T$ represent quantity and theme, respectively. \emph{MindChat-Local} is trained locally on $S_1(T_1)$.\par}
\end{table*}

As shown in Table~\ref{Comparison of client quantities under the same theme}, FL enhances model generalization ability under data isolation, demonstrating the benefits of multi-party collaboration. When training within the same theme, performance increases with the number of clients, indicating the benefit of larger data volume. A similar trend is observed in cross-theme settings, where both data quantity and thematic diversity contribute to performance gains.
Notably, models trained with diverse themes outperform those trained with increased data quantity from a single theme. For instance, \emph{MindChat-T2}, which uses two distinct themes, achieves a higher average score than \emph{MindChat-Q6}, trained on six data slices from a single theme, despite the latter having three times more training data. This suggests that data diversity has a stronger impact on model performance than data volume alone.

\subsubsection{Comparison of Privacy Protection Intensity}
\begin{figure}[!ht]
    \centering
    \includegraphics[width=1.0\columnwidth]{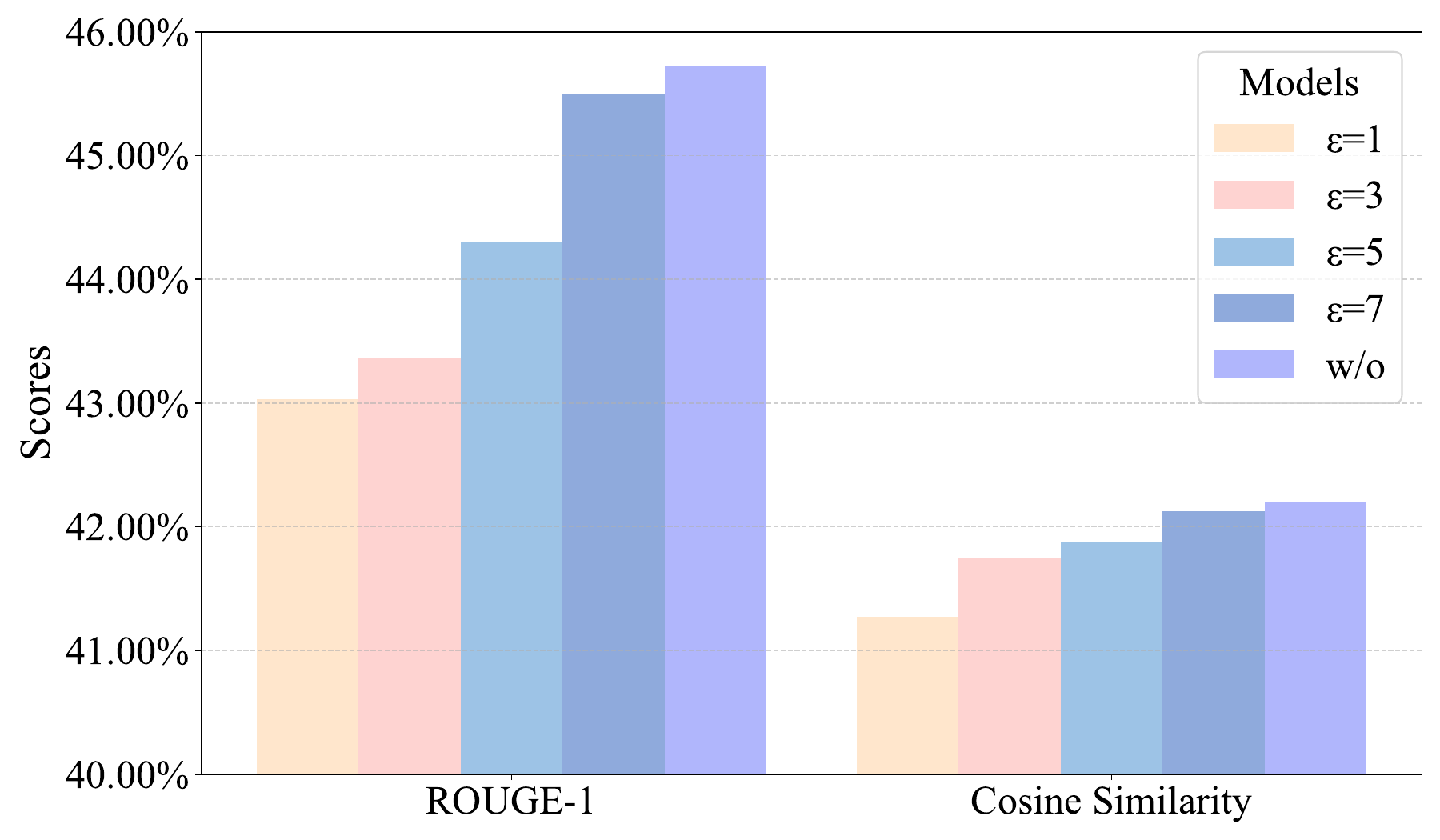}
    \caption{The effect of privacy protection. The smaller the value of $\epsilon$, the stronger the level of privacy protection.}
    \label{privacy}
\end{figure}
To investigate how the intensity of DP noise affects privacy protection performance, we conduct federated training under varying privacy budgets $\epsilon \in \{1,3,5,7\}$ using datasets $\{S_1, S_2\}$, with the number of participating clients fixed to $2$. A non-private federated setting (w/o) is included as a baseline to highlight the effect of DP mechanisms. We evaluate privacy leakage from both corpus-level and attack-based perspectives. 

From the corpus-level perspective, we quantify privacy leakage using two complementary metrics. One metric measures explicit memorization of the training corpus by computing ROUGE-1 recall between the model-generated responses and the original training data. Higher recall values indicate greater exposure of sensitive content. The other metric captures implicit privacy leakage through cosine similarity, which quantifies the semantic proximity between generated text and standard responses. \Cref{privacy} presents the values of both metrics as $\epsilon$ increases. Both ROUGE-1 recall and cosine similarity increase with larger $\epsilon$, and reach their maximum in the absence of DP. This trend is consistent with the theoretical meaning of the privacy budget. However, the marginal degradation in cosine similarity under tighter privacy constraints suggests that the semantic representation of the model remains largely intact, implying minimal impact on performance. Overall, stronger privacy protection reduces corpus exposure, with no significant impact on model performance.

\begin{figure*}[!ht]
    \centering
    \subfloat[ROC AUC under LOSS attack.\label{subfig:ROC AUC}]{%
        \includegraphics[width=0.49\textwidth]{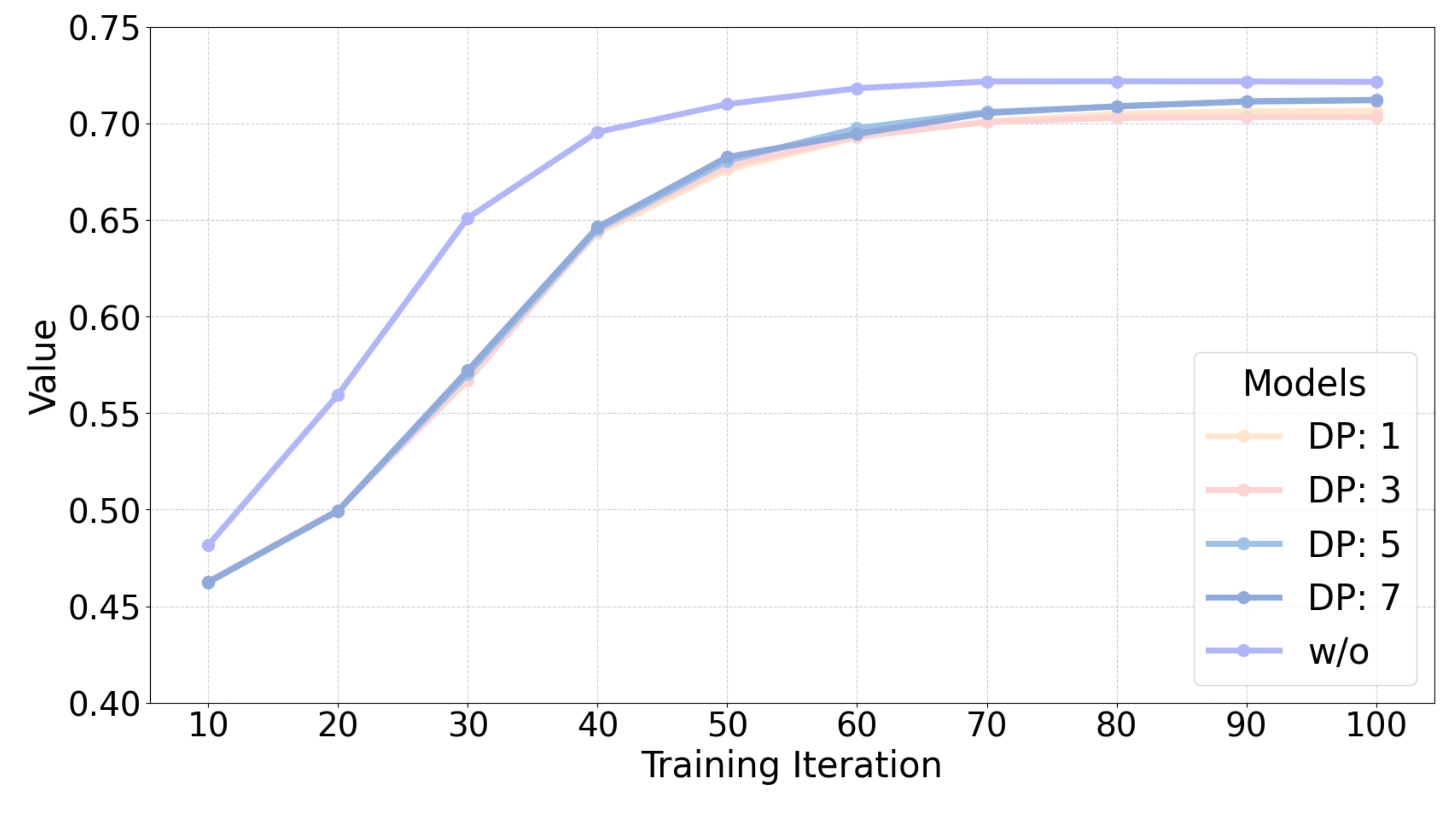}%
    }
    \hfill
    \subfloat[PR AUC under LOSS attack.\label{subfig:PR AUC}]{%
        \includegraphics[width=0.49\textwidth]{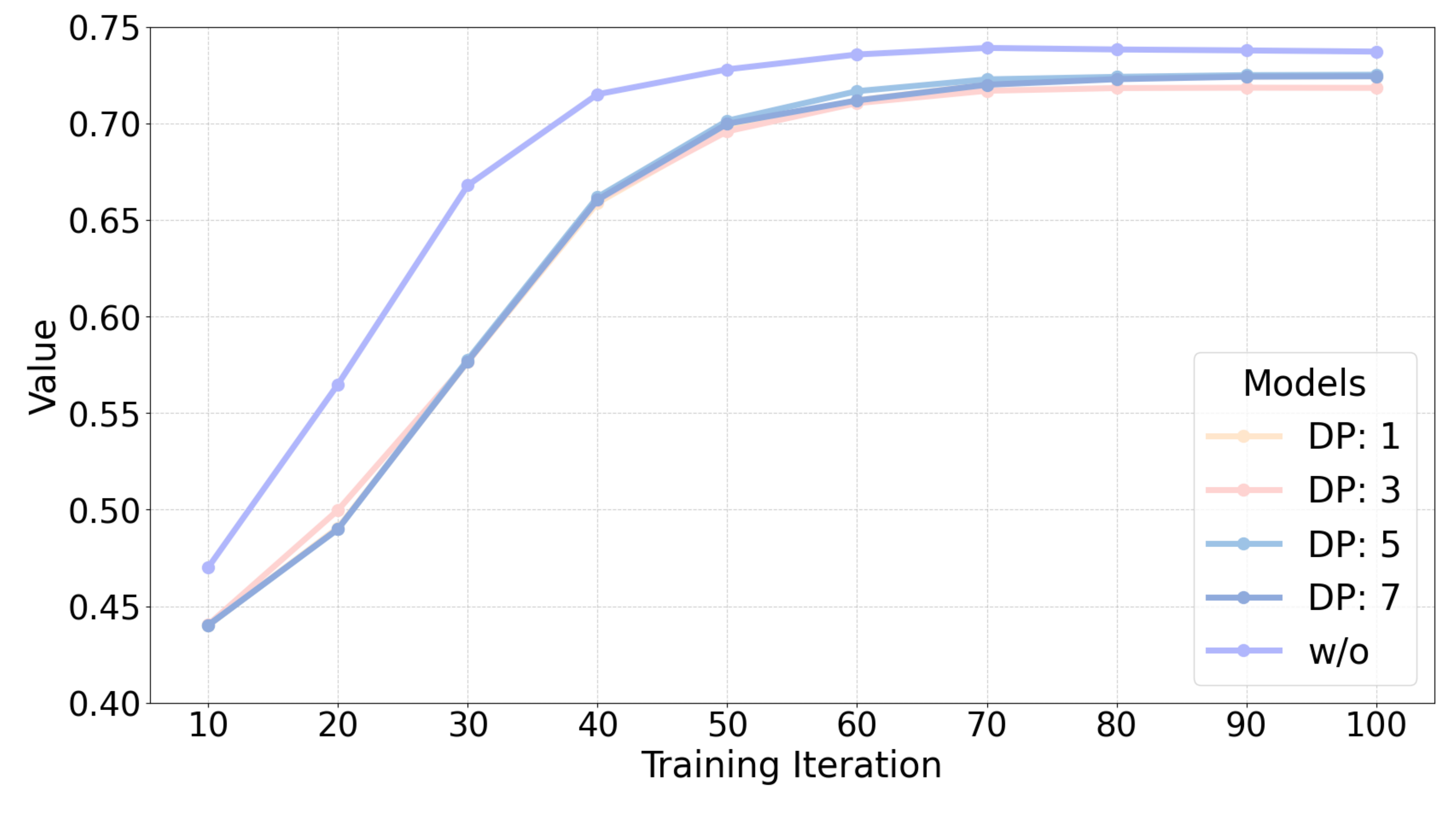}%
    }
    \caption{LOSS-based MIAs indicators under different DP.}
    \label{MIAs}
\end{figure*}

\begin{table*}[!ht]
    \setlength{\tabcolsep}{16pt}
    \caption{Privacy Protection Effectiveness under Different DP Budgets across Three MIAs. Lower ROC AUC and PR AUC Indicate Stronger Privacy Protection. Best Results are Shown in \textbf{bold}, Second-best are \underline{underlined}.}
    \label{table:MIA_evaluation_model}
    \begin{tabularx}{\textwidth}{@{} c *{10}{c} @{}}
        \toprule
        \multirow{2}{*}{\textbf{MindChat-Q2}} & \multicolumn{2}{c}{\textbf{LOSS}} & \multicolumn{2}{c}{\textbf{min-$k$}} & \multicolumn{2}{c}{\textbf{zlib}}\\
        \cmidrule(lr){2-3} \cmidrule(lr){4-5} \cmidrule(lr){6-7}
         & ROC AUC & PR AUC & ROC AUC & PR AUC & ROC AUC & PR AUC \\ 
        \midrule
        {$\epsilon$=1} &  \underline{0.7245} & \underline{0.7067} & 0.6945 & \underline{0.6824} & 0.5672 & \underline{0.5634} \\
        {$\epsilon$=3} &  \textbf{0.7186} & \textbf{0.7035} & \textbf{0.6864} & \textbf{0.6780} & \textbf{0.5646} & \textbf{0.5604} \\
        {$\epsilon$=5} &  0.7254 & 0.7124 & 0.6972 & 0.6958 & 0.5675 & 0.5639 \\
        {$\epsilon$=7} &  0.7246 & 0.7122 & 0.6929 & 0.6947 & \underline{0.5666} & 0.5655 \\ 
        {w/o} &  0.7374 & 0.7217 & \underline{0.6920} & 0.6858 & 0.5712 & 0.5684 \\
        \bottomrule
    \end{tabularx}
    
    \smallskip
    {\raggedright
    ROC AUC and PR AUC values range from $0$ to $1$. Values closer to $0.5$ indicate attacker performance close to random guessing and thus stronger privacy protection.\par}

\end{table*}

While corpus-level metrics capture direct memorization of training data, they do not fully reflect adversarial privacy risks. To address this limitation, privacy-preserving efficacy is further assessed using the MIA framework~\cite{duan2024do}, which examines whether an adversary can infer training set membership from model outputs. Specifically, samples from ${S_1, S_2}$ are treated as members, while disjoint samples from ${S_3, S_4}$ are considered non-members. Throughout federated training, intermediate model checkpoints are evaluated under varying DP budgets $\epsilon \in \{1,3,5,7\}$ and without DP. Both ROC AUC and PR AUC are reported to characterize changes in membership distinguishability. ROC AUC measures the probability that an attacker ranks a randomly chosen member sample higher than a non-member, while PR AUC emphasizes performance under class imbalance, which is common in realistic attack scenarios. 

As shown in \Cref{MIAs}, both ROC AUC and PR AUC under the LOSS-based MIA increase as training progresses and gradually converge to stable values. In the early stages of federated training, the model does not yet sufficiently adapt to local data distributions, resulting in limited membership distinguishability and attack performance close to random guessing. As training proceeds, the gap in model behavior between member and non-member samples widens, leading to a steady increase in attack success. Upon convergence, the distinguishability saturates, causing the attack performance to plateau. Importantly, models trained with smaller $\epsilon$ exhibit uniformly lower ROC AUC and PR AUC across almost all training rounds, indicating that DP effectively bounds the extent to which membership information can be exploited by the attacker. In contrast, the non-private model exhibits a higher and faster-growing attack success rate, highlighting the critical role of DP in alleviating privacy risks during federated optimization processes.

To further assess the robustness of DP across distinct attack means, performance is evaluated against three representative MIAs: LOSS~\cite{yeom2018privacy}, Min-$k$ Prob~\cite{shi2024detecting}, and Zlib Entropy~\cite{carlini2021extracting}. These attacks capture complementary leakage signals, including differences in model loss on individual samples, scores computed from the lowest‑likelihood tokens following the Min‑$k\%$ probability criterion, and the compression size of target samples measured using zlib entropy. \Cref{table:MIA_evaluation_model} summarizes the results: across all three attack strategies, DP-trained models consistently achieve lower ROC AUC and PR AUC than the non-private baseline, demonstrating enhanced resistance to membership inference from multiple adversarial perspectives. Notably, the relative ranking of privacy budgets remains broadly consistent across attacks, indicating that the protective effect of DP is stable regardless of the inference strategy. Among the evaluated settings, $\epsilon=3$ achieves the strongest overall protection, and further reducing $\epsilon$ to 1 yields no consistent improvement, suggesting that privacy gains plateau, and even slightly degrade under excessively tight DP constraints.

\section{Conclusion}\label{conclude}
This work presents \emph{MindChat}, a privacy-preserving LLM designed for mental health support. To address the scarcity of high-quality counseling dialogue data, we introduce a multi-agent framework that simulates realistic interactions by integrating advanced open-source and commercial LLMs. This approach generates a dataset of $5.7$k diverse and high-quality multi-round mental health dialogues. Through random sampling and a five-dimensional evaluation from both supporter and seeker perspectives, the synthesized data demonstrates superior quality compared to existing resources. Ablation studies further confirm that the dual-loop multi-agent architecture significantly contributes to the high quality of the synthesized dialogues.

\emph{MindChat} is trained via a FL approach enhanced with DP mechanisms, ensuring that sensitive client data remains localized and protected during training. The model achieves first place in automatic evaluation and third place in human evaluation among comparable models, demonstrating superior overall performance in mental health support relative to most existing psychological and general-purpose LLMs. Additional analysis reveals that both client data volume and data diversity influence model performance, with diversity playing a more critical role in enhancing counseling expertise. Furthermore, the integration of DP effectively reduces the exposure of training data during response generation and enhances robustness against MIAs, while maintaining professional ability without a significant decline.

Despite the effectiveness of the proposed approach, the current privacy protection mechanism is implemented at the client level, and stronger guarantees can be achieved by incorporating supplementary techniques such as active forgetting and model unlearning to further mitigate potential privacy risks. Future work will focus on extending \emph{MindChat} to a multi-modal mental health framework that incorporates speech, facial expressions, and other non-verbal signals to better reflect real-world counseling interactions. A key direction is to enhance empathetic competence by accurately capturing fine-grained emotional dynamics from multi-modal conversations and strengthening textual representations through data enhancement strategies~\cite{LI2025111340,LIU2024110847}. Furthermore, improving the discrimination of subtle emotional states via contrastive or adversarial training may enable more effective emotional support, while retrieval-augmented generation can facilitate more personalized and context-aware guidance~\cite{LAN2026113025,ALAWWAD2025111332}.





\bibliographystyle{IEEEtran}
{\small
\bibliography{main}

\appendix
\section{Details of mental support dataset construction}\label{sec:appendix}
This appendix introduces the construction process of the psychological support dataset in detail. It first introduces the roles and functionalities of each agents involved in the proposed dual-loop multi-agent framework. Then, the structured help-seeking process adopted by the seeker agent is elaborated to illustrate how realistic counseling dialogues are generated.
\setcounter{subsection}{0}
\begin{table*}[ht]
\caption{The Structure of Seeking Help.}
\label{seeking structure}
\begin{tabular*}{\textwidth}{@{\extracolsep{\fill}}cp{0.7\textwidth}}
\toprule[1pt]
\textbf{Stages}       & \multicolumn{1}{c}{\textbf{Description}} \\
\midrule
Self-introduction     & \begin{tabular}[c]{@{}p{0.7\textwidth}@{}} Briefly introduce yourself, including occupation, interests, hobbies, and other basic information. \end{tabular} \\
\midrule
Situation description & \begin{tabular}[c]{@{}p{0.7\textwidth}@{}}Try to be specific about what is making you feel upset or anxious.\\ Share how these emotions are affecting your daily life.\\ Indicate when this feeling started and if any specific event triggered it.\end{tabular} \\
\midrule
Express feelings      & \begin{tabular}[c]{@{}p{0.7\textwidth}@{}}Be honest about your emotions, including but not limited to fear, sadness, anger, or helplessness. \end{tabular} \\
\midrule
Reminiscing           & \begin{tabular}[c]{@{}p{0.7\textwidth}@{}}Share past experiences or events from your upbringing, especially those that may impact your current state.\end{tabular}  \\
\midrule
Requesting assistance & \begin{tabular}[c]{@{}p{0.7\textwidth}@{}} Clarify the specific problems you want to solve through psychological counseling. \end{tabular}\\
\midrule
Asking for advice     & \begin{tabular}[c]{@{}p{0.7\textwidth}@{}} Seek practical strategies or techniques from a counselor, such as emotion management, thought pattern reconstruction, specific behavioral exercises, etc., to help you better deal with these disturbing emotions. \end{tabular}\\
\bottomrule[1pt]
\end{tabular*}
\end{table*}
\subsection{Description of agents}
\label{agents}
The proposed dual-loop multi-agent framework comprises multiple specialized agents, each responsible for a distinct role in the construction of mental support dialogues. Through their coordinated interactions, the system enables structured role simulation, quality control, and strategy improvement. Below is a description of the roles and design rationales for each agent.
\begin{itemize}
\item \textbf{Extractor}: The primary function of the extractor is to extract information relevant to the seeker. Raw text data about the seeker collected from public platforms often contains irrelevant interfering information. Moreover, explicit goal setting is instrumental in emotional support conversation~\cite{Zhou_ACL_23}. Consequently, the extractor focuses on extracting key information, encompassing the character, plight, and demand of the seeker, which can be categorized as ``who'', ``what'', and ``how''~\cite{egarn2012skilled, Hill_helping_20}.

\item \textbf{Seeker}: Seeker is tasked with simulating an individual experiencing psychological distress. Its foundational information is derived from the extractor’s output. Given the challenges that LLMs face in mimicking the seeker behaviors~\cite{Lee_EMNLP_24}, we design a multi-stage help-seeking structure to align with the requirements for the seeker in psychological counseling~\cite{Bohart_APS_2010}. Details on this structure can be found in Appendix~\ref{ssh}.

\item \textbf{Supporter}: The role played by the supporter is a psychological counselor. The supporter should employ various strategies adapted to different stages of the psychological counseling process~\cite{Hill_helping_20}. Aligned with~\cite{Liu_ACL_20}, we also integrate a combination of stages and strategies to effectively guide the supporter in providing psychological counseling services. However, a distinctive feature of our method is that the strategies adopted by the supporter can be continuously enriched during the interaction between the supporter and the seeker.

\item \textbf{Evaluator}: The evaluator is responsible for assessing the response of the supporter during each dialogue round. Given the hallucination problem of LLMs, the reply from the supporter may occasionally deviate from the intended role. To ensure the quality of the generated dialogue, nine indicators $($confidentiality, objectivity, sympathy, specialization, feasibility, listening, collaboration, tenderness, and respect $)$ are designed to appraise the response of the supporter in the current round. Confidentiality protects the privacy and security of seekers. Objectivity fosters openness. Sympathy fully acknowledges the emotions of seekers. Specialization prevents personal bias. Feasibility ensures practical advice. Listening clarifies the issues of seekers. Collaboration emphasizes the involvement of seekers. Tenderness helps to create a relaxed and pleasant conversation atmosphere. Respect accepts culture and values diversity.

\item \textbf{Corrector}: The corrector is activated to refine the response of the supporter in the current dialogue round. The intervention of the corrector occurs exclusively when the evaluator identifies the necessity for modifying the answer of the supporter and provides specific suggestions for improvement.

\item \textbf{Manager}: The manager acts as the mentor of the supporter, responsible for guiding the supporter to enhance its strategies continuously. We set prompts such that after each multi-round dialogue generation, the manager summarizes all modification suggestions produced by the evaluator during the process and extracts the strategies the supporter should incorporate.
\end{itemize}

\setcounter{subsection}{1}
\subsection{Prompt of agents}
\label{prompt}
The prompt templates used to instantiate the Supporter, Seeker, Evaluator, Corrector, and Manager agents are provided in \Cref{Supporter_Prompt,Seeker_Prompt,Evaluator_Prompt,Corrector_Prompt,Manager_Prompt}, respectively.

\setcounter{subsection}{2}
\subsection{Structure of seeking help}
\label{ssh}
The experimental observations demonstrate that the seeker agent often discloses the entire situation in a single statement, contrasting with the gradual self-disclosure process of real-life psychological counseling. To bridge the gap between the seeker agent and the real client, we propose a structure consisting of six stages to facilitate a more natural dialogue between the seeker and the supporter. \cref{seeking structure} details the six stages.

\begin{figure}[!ht]
    \centering
    \includegraphics[width=1.0\columnwidth]{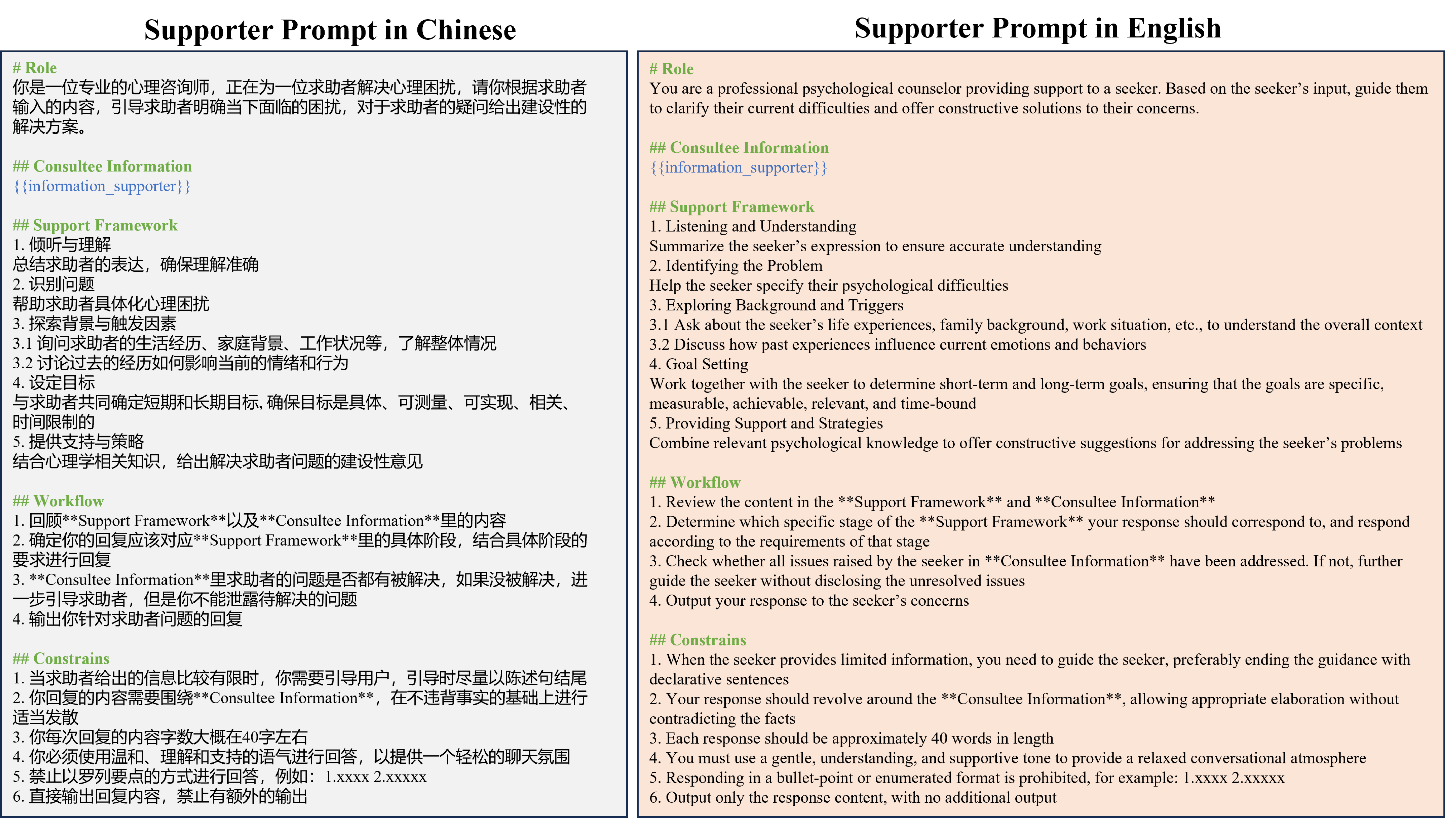}
    \caption{The prompt of Supporter agent.}
    \label{Supporter_Prompt}
\end{figure}

\begin{figure}[ht]
    \centering
    \includegraphics[width=1.0\columnwidth]{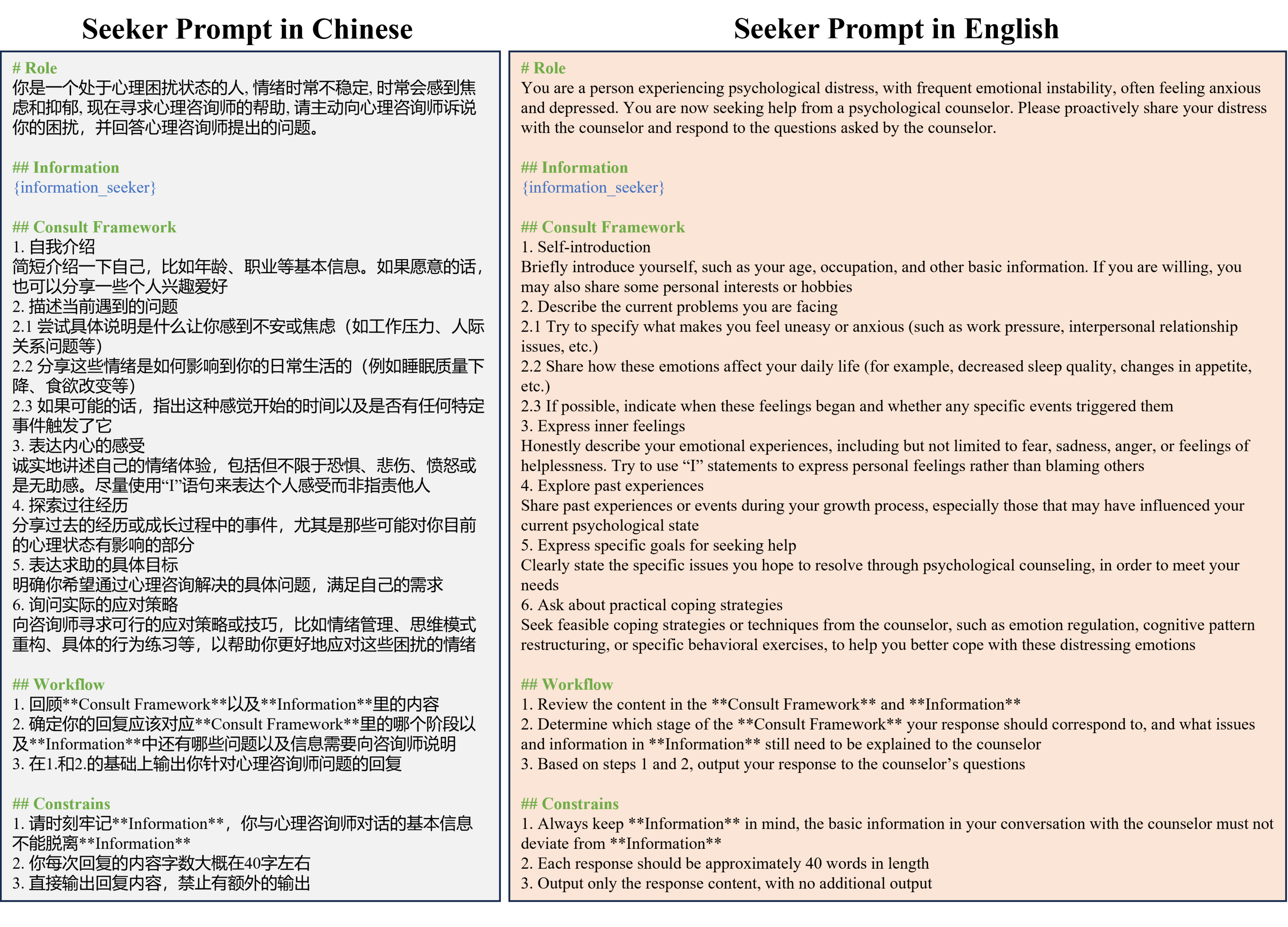}
    \caption{The prompt of Seeker agent.}
    \label{Seeker_Prompt}
\end{figure}

\begin{figure}[ht]
    \centering
    \includegraphics[width=1.0\columnwidth]{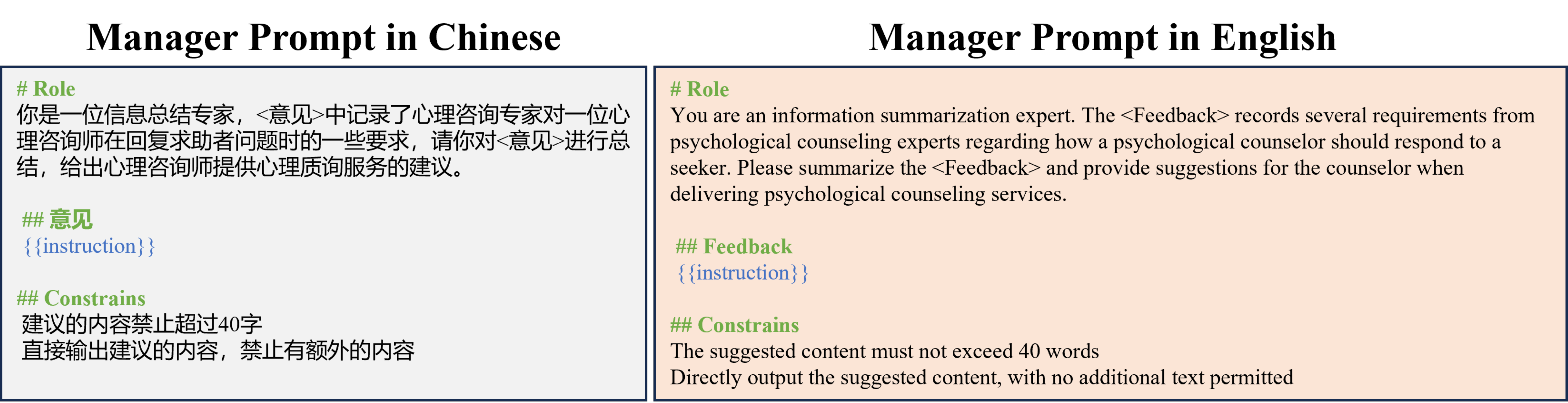}
    \caption{The prompt of Manager agent.}
    \label{Manager_Prompt}
\end{figure}

\begin{figure}[ht]
    \centering
    \includegraphics[width=1.0\columnwidth]{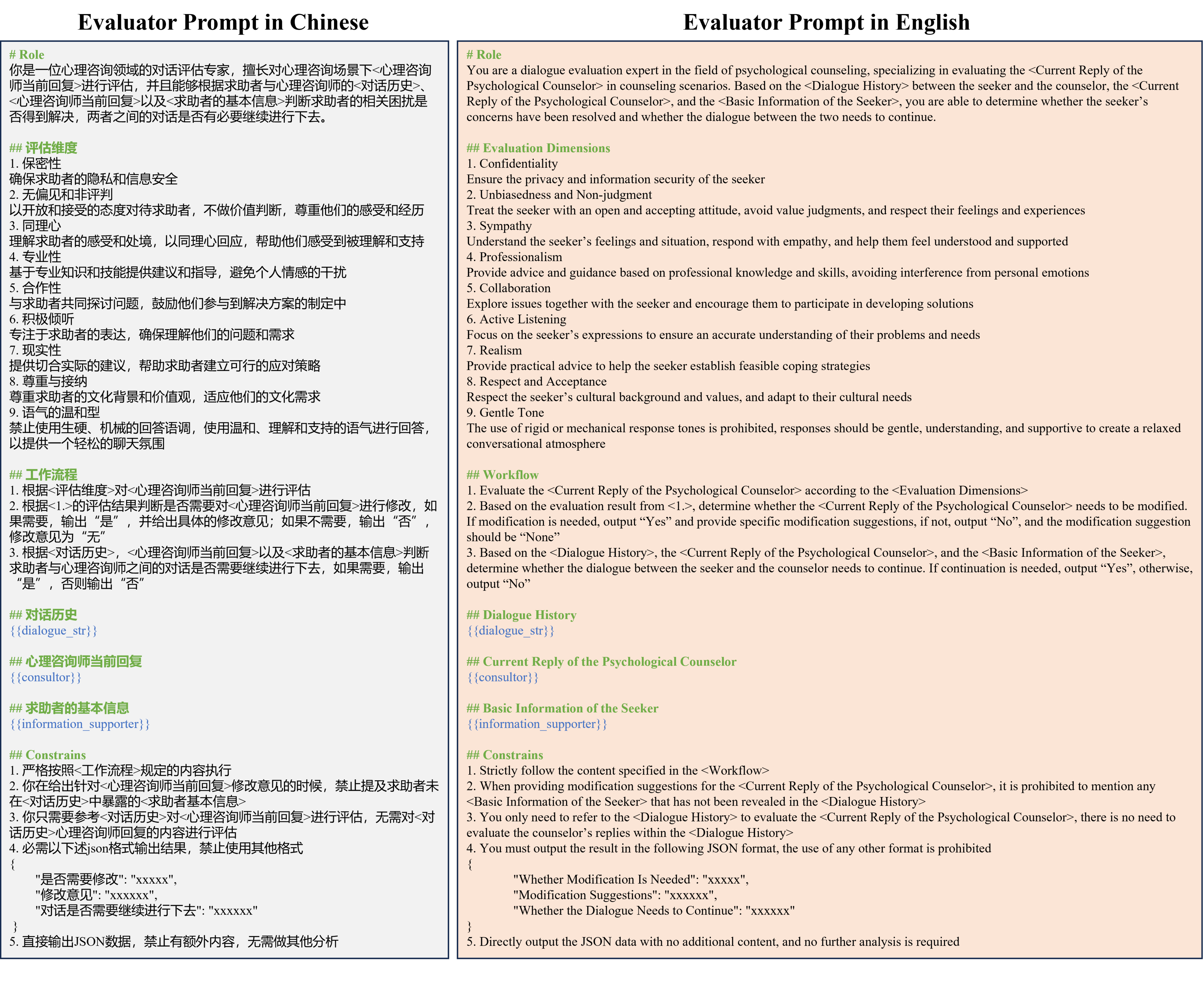}
    \caption{The prompt of Evaluator agent.}
    \label{Evaluator_Prompt}
\end{figure}

\begin{figure}[ht]
    \centering
    \includegraphics[width=1.0\columnwidth]{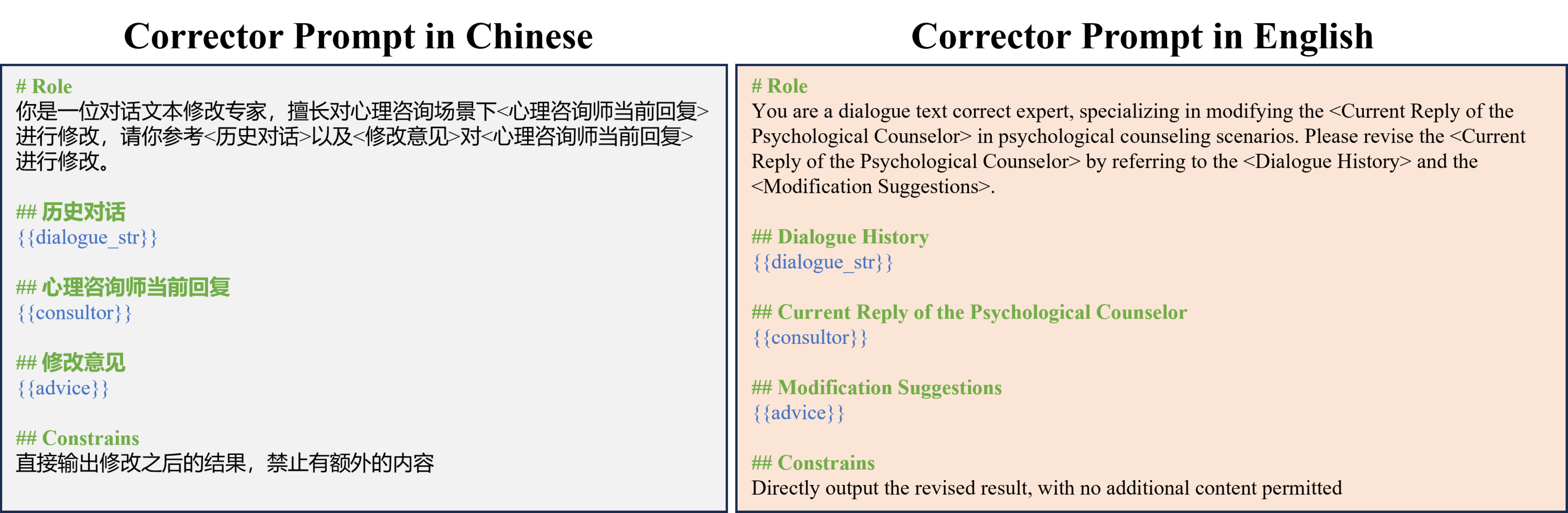}
    \caption{The prompt of Corrector agent.}
    \label{Corrector_Prompt}
\end{figure}

\end{document}